\documentclass[accepted]{includes/uai2026} 

\usepackage[british]{babel}

\usepackage{amsmath,amsfonts,bm}

\newcommand{\dd}{\mathrm{d}}

\def\1{\bm{1}}

\DeclareMathAlphabet{\mathsfit}{\encodingdefault}{\sfdefault}{m}{sl}
\SetMathAlphabet{\mathsfit}{bold}{\encodingdefault}{\sfdefault}{bx}{n}

\newcommand{\E}{\mathbb{E}}

\newcommand{\R}{\mathbb{R}}

\usepackage{xcolor}
\definecolor{astral}        {RGB}{46,116,181}
\definecolor{cb-blue}       {RGB}{70, 130, 180}
\definecolor{orange}        {RGB}{214,150, 92}
\definecolor{green}         {RGB}{136,196,136}
\usepackage{hyperref}
\usepackage{cleveref}
\hypersetup{colorlinks,linkcolor={cb-blue},citecolor={cb-blue},urlcolor={cb-blue}}
\usepackage{url}
\usepackage{booktabs}
\usepackage{multirow}
\usepackage{multicol}
\usepackage{subcaption}
\usepackage{algorithm}
\usepackage{algpseudocode}
\usepackage{graphicx}
\usepackage{amssymb}
\usepackage{amsmath}
\usepackage{amsthm}
\usepackage{mathtools}
\usepackage{minitoc}
\usepackage{cleveref}
\usepackage{wrapfig}

 \usepackage[textwidth=1.8cm, textsize=scriptsize]{todonotes}

\usepackage[normalem]{ulem}

\usepackage[most]{tcolorbox}

\newtcolorbox{codexbox}{
  breakable,
  colback=orange!10,
  colframe=orange!30!red,
  boxrule=0.4pt,
  arc=1mm,
  left=1mm,
  right=1mm,
  top=1mm,
  bottom=1mm,
  fonttitle=\bfseries,
  title=codex
}

\newtcolorbox{denizbox}{
  breakable,
  colback=red!10,
  colframe=red!60!black,
  boxrule=0.4pt,
  arc=1mm,
  left=1mm,
  right=1mm,
  top=1mm,
  bottom=1mm,
  fonttitle=\bfseries,
  title=\"{o}da
}

\newtcolorbox{claudebox}{
  breakable,
  colback=violet!8,
  colframe=violet!60!black,
  boxrule=0.4pt,
  arc=1mm,
  left=1mm,
  right=1mm,
  top=1mm,
  bottom=1mm,
  fonttitle=\bfseries,
  title=claude
}

\definecolor{mypink}{rgb}{0.85,0.1,0.4}

\algrenewcommand\algorithmiccomment[1]{\hfill\textcolor{mypink}{\texttt{\# #1}}}

\newtheorem{theorem}{Theorem}

\newtheorem{proposition}{Proposition}

\newtheorem{assumption}{\textbf{A}\hspace{-0.12cm}}
\Crefname{assumption}{\textbf{A}\hspace{-0.12cm}}{\textbf{H}\hspace{-3pt}}
\crefname{assumption}{\textbf{A}}{\textbf{A}}

\theoremstyle{definition}
\newtheorem{remark}{Remark}
\makeatletter
\let\oldremark\remark
\let\endoldremark\endremark
\renewenvironment{remark}{\oldremark}{\hfill$\diamond$\endoldremark}
\makeatother

\usepackage[acronym]{glossaries}
\makeglossaries
\newacronym[plural=EBMs, firstplural=Energy-Based Models (EBMs)]{ebm}{EBM}{Energy-Based Model}
\newacronym[plural=LEBMs,firstplural=Latent Energy-Based Models (LEBMs)]{lebm}{LEBM}{Latent Energy-Based Model}

\newacronym{mmd}{MMD}{Maximum Mean Discrepancy}
\newacronym{iplebm}{EBIPLA}{Energy-Based Interacting Particle Langevin Algorithm}
\newacronym{mle}{MLE}{Maximum Likelihood Estimation}
\newacronym{mmle}{MMLE}{Maximum Marginal Likelihood Estimation}
\newacronym{mcmc}{MCMC}{Markov chain Monte Carlo}
\newacronym{ula}{ULA}{Unadjusted Langevin Algorithm}
\newacronym{ipla}{IPLA}{Interacting Particle Langevin Algorithm}
\newacronym{pgd}{PGD}{Particle Gradient Descent}
\newacronym{em}{EM}{Expectation-Maximization}
\newacronym{smc}{SMC}{sequential Monte Carlo}
\newacronym{lvm}{LVM}{Latent Variable Model}
\newacronym{FID}{{{\textsc{\small FID}}}}{Fréchet Inception Distance}

\newcommand{\ebm}{\gls*{ebm}}
\newcommand{\ebms}{\glspl*{ebm}}

\usepackage[round]{natbib}

\usepackage{subfiles}

\title{Learning Latent Energy-Based Models via Interacting Particle Langevin Dynamics}

\author[*]{Joanna~Marks}
\author[*]{Tim~Y.~J.~Wang}
\author[ ]{O.~Deniz~Akyildiz}
\affil[ ]{%
    Department of Mathematics, Imperial College London
}

\begin{document}
\maketitle

\begin{abstract}
We develop interacting particle algorithms for learning latent variable models with energy-based priors. To do so, we leverage recent developments in particle-based methods for solving maximum marginal likelihood estimation (MMLE) problems. Specifically, we provide a continuous-time framework for learning latent energy-based models, by defining stochastic differential equations (SDEs) that provably solve the MMLE problem. We obtain a practical algorithm as a discretisation of these SDEs and provide theoretical guarantees for the convergence of the proposed algorithm. Finally, we empirically validate the  effectiveness of our method on synthetic and image datasets and demonstrate that using a particle based approach offers significant improvement in computational efficiency.
\end{abstract}

\section{INTRODUCTION}

\glspl*{ebm} offer a flexible framework for statistical and generative modelling by learning models of the form $p_\theta \propto e^{-U_\theta}$ from data. The energy function $U_\theta$ can be flexibly parameterised, enabling \glspl*{ebm} to approximate any probability density \citep{atchade2023generative}. Once the parameters $\theta$ are learned for a given dataset, these models support sampling and inference, allowing their use in a wide range of applications, including generation across modalities \citep{deng2020residualenergybasedmodelstext, gladstone2025}, robust classification \citep{grathwohl2020classifier}, anomaly detection \citep{zhai2016}, likelihood-based evaluation \citep{du2019}, simulation-based inference \citep{glaser2022maximum}, and compositional generation \citep{du2020compositional,du2023reduce}.

\gls*{mle} procedures to learn \glspl*{ebm} remain highly challenging to implement since common methods require samples from the model $p_\theta$ to perform parameter updates, typically drawn using \gls*{mcmc} methods \citep{hinton2002training,song2021train}. In high dimensional settings, these often suffer from slow mixing times. Moreover, real-world data often concentrates on a low-dimensional manifold \citep{bengio2014representationlearningreviewnew,whiteley2025statistical}, meaning that the learning task might be ill-defined \citep{loaiza2024deep}. 

A natural way to address both challenges within the \ebm~framework is to learn a low-dimensional latent representation of the data and model the encoded data as an \gls*{ebm}, giving rise to \gls*{lebm} \citep{pang2020ebmprior}. This model comprises of a likelihood (decoder) and a prior in the (lower-dimensional) latent space, which is parameterised as an \gls*{ebm}. {The latent formulation, however, introduces new challenges} in the \gls*{mle}-training of such models. First, the model parameters must be learned by maximising the \textit{marginal likelihood} of the observed data, which involves integrating out the latent variables, a typical setting known as \gls*{mmle} \citep{dempster1977maximum}. Second, the normalising constant of the prior distribution is generally intractable, resulting in a ``doubly-intractable'' setting.

 A notable method to tackle this problem is built by \citet{pang2020ebmprior}, who propose an \gls*{mcmc}-based approach to tackle both the intractable posterior and the intractable normalising constant of the prior. This procedure can be seen as an \gls*{em}-like procedure, in a similar spirit to \citet{debortoli2021soul}. Adapted to the \ebm~setting, these methods involve (double) \gls*{mcmc} loops to sample from the posterior and/or the prior at each iteration of the training, which introduce significant computational overhead and complicate the non-asymptotic analysis.

In this work, we introduce a novel, diffusion-based approach to solve the \gls*{mmle} problem for the \ebm~prior latent variable model. By leveraging the rich connection between Langevin dynamics and optimisation \citep{raginsky2017non, zhang2023nonasymptotic}, we develop a training algorithm based on an interacting particle system that simultaneously estimates the parameters and the posterior distribution through solving a system of SDEs.

\noindent\textbf{Contributions.} Our contributions can be summarised as follows:
\begin{itemize}
\item In Section~\ref{sec:algorithm}, we introduce a continuous-time, SDE-based framework for training latent~\ebms. This interacting particle system is inspired by the \gls*{ipla} \citep{akyildiz2023ipla} and adapted to the \gls*{lebm}~setting. By construction, the $\theta$-marginal of the stationary measure of the proposed system concentrates on the maximisers of the marginal likelihood with increasing $N$ under very mild assumptions. We then develop an algorithm by discretising SDE systems that target the \gls*{mmle} solution, termed \gls*{iplebm}, given in Algorithm~\ref{algo:simple}.
\item Section~\ref{sec:nonasymptotic} provides a non-asymptotic analysis for our method under log-concavity and smoothness assumptions. We show that, under these assumptions and for a suitable discretisation step size, the parameter estimates $(\theta_k)_{k\geq 0}$ generated by \gls*{iplebm} approach the empirical maximiser $\theta_\star$ as the iterations $k$ and number of particles $N$ increase.
These results constitute the first convergence bounds for training \glspl*{lebm}.

\item Furthermore, by introducing a novel rescaling, we derive convergence bounds that explicitly depend on the number of data points $M$ and show that the distance of the parameter estimates to the empirical maximiser $\theta_\star$ decreases as $M$ increases, which to the best of our knowledge has not been addressed in the available \gls*{ipla} literature. This provides a theoretical justification for why particle-based methods remain effective even when using a relatively small number of particles on large datasets. 
\item In Section~\ref{sec:experiments}, we empirically validate the effectiveness of our method on synthetic datasets and image generation tasks, showing that it achieves competitive performance among relevant baselines. We also demonstrate that using a particle based approach offers significant improvement in computational efficiency.
\end{itemize}
\noindent\textbf{Notation.} We use $\mathcal{O}(\cdot)$ to denote upper bounds up to constant factors, and $\widetilde{\mathcal{O}}(\cdot)$ to denote the same while suppressing polylogarithmic terms. For a random variable $X$, the notation $\mathcal{L}(X)$ denotes the law (distribution) of $X$ while $\sigma(X)$ denotes the $\sigma$-algebra generated by $X$. For any two probability distributions $\mu$ and $\nu$, we define the Wasserstein distance:
\begin{equation*}
    W_2(\mu,\nu) \coloneqq \left(\inf_{\gamma \in \Gamma(\mu,\nu)} \int \|x-y\|^2 \, \mathrm{d} \gamma(x,y)\right)^{1/2},
\end{equation*}
where $\Gamma(\mu,\nu)$ is the set of all couplings of $\mu$ and $\nu$. The notation $\delta_x$ denotes the Dirac measure at $x$. Where appropriate, we use the notation $x^{1:M} = \{x^m\}_{m=1}^M$ and $x^{1:M,1:N} = \{x^{m,n}\}_{m=1,n=1}^{M,N}$ to denote collections of variables.

\section{BACKGROUND}

We start by defining a real-valued latent energy-based probabilistic model \citep{pang2020ebmprior,yu2023diffusionamortized}:
\begin{equation}\label{eq:latent-ebm-model}
p_{\theta}(x, y) = p_{\alpha}(x)  p_\beta(y \mid x),
\end{equation}
where $\theta = (\alpha, \beta) \in \R^{d_\alpha + d_\beta}  =\R^{d_\theta}$  and $p_{\beta}(y | x) = \mathcal{N}\big(y; g_{\beta}(x), \sigma^2 I \big)$ is an isotropic Gaussian distribution. Here, $g_{\beta}: \R^{d_x} \to \R^{d_y}$ denotes a generator function parametrised by $\beta$ that maps from the latent space to the ambient space.
The prior $p_{\alpha}(x)$ is modelled as an \gls*{ebm}:
\begin{equation}
p_\alpha(x) = \frac{1}{Z(\alpha)} e^{-U_\alpha(x)},
\end{equation}
where $Z(\alpha) = \int e^{-U_\alpha(x)} \mathrm{d}x$ is the $\alpha$-dependent normalising constant and we assume $Z(\alpha)<\infty$ for all $\alpha\in \R^{d_\alpha}$ to ensure the model is well defined.
When $U_{\alpha}(x)$ is chosen to be a neural network, the \gls*{ebm} can capture meaningful latent structure.

\subsection{Maximum marginal likelihood estimation}\label{sec:mmle}
Given a dataset $\{y_m\}_{m=1}^M \sim p_{\text{data}}$, we are interested in learning the parameters $\theta$ of the model by maximising the marginal likelihood of the observed data. This problem is termed the \gls*{mmle} problem. Specifically, we want to solve
\begin{align*}
\theta_\star^{\text{pop}} \in \arg\max_{\theta \in \mathbb{R}^{d_\theta}} \ell(\theta),
\qquad 
\ell(\theta) := \mathbb{E}_{p_{\text{data}}}\big[\log p_\theta(Y)\big].
\end{align*}
where $p_\theta(y) = \int p_\alpha(x)p_\beta(y|x)  \mathrm{d}x.$

Since we do not have access to the true data distribution $p_{\text{data}}$, we can approximate the expectation over $p_{\text{data}}$ by an empirical measure ${p}^M_{\text{data}} = ({1}/{M})\sum_{m=1}^M \delta_{y_m}(\mathrm{d} y)$ which yields the empirical loss function 
\begin{equation}
    \ell_M(\theta)=\frac{1}{M} \sum_{m=1}^M \log p_\theta(y_m)
\end{equation}
converting the population \gls*{mmle} problem to the empirical \gls*{mmle} problem:
\begin{align}\label{eq:empirical-marginal-loglikelihood}
\theta_\star \in \arg\max_{\theta \in \mathbb{R}^{d_\theta}} \frac{1}{M} \sum_{m=1}^M \log p_\theta(y_m),
\end{align}
where $M$ denotes the number of data-points. Our main goal in this paper is to solve \eqref{eq:empirical-marginal-loglikelihood} efficiently and to provide theoretical guarantees for the obtained solution.

\subsection{Optimisation via Langevin dynamics}

Following \citet{akyildiz2023ipla}, in order to maximise $\ell(\theta)$, we will exploit stochastic dynamics whose long-run behaviour naturally concentrates around the maximisers. A classical choice is \emph{Langevin dynamics}, given by the SDE \citep{raginsky2017non,zhang2023nonasymptotic}
\begin{align}\label{eq:langevin_dynamics_for_optimisation}
\mathrm{d}\theta_t = \nabla_\theta \ell(\theta_t)\,\mathrm{d}t 
+ \sqrt{{2 / \eta}}\,\mathrm{d}B_t,
\end{align}
where $(B_t)_{t \geq 0}$ is a standard $d_\theta$-dimensional Brownian motion and $\eta > 0$ is an inverse temperature parameter. Under standard regularity assumptions, this SDE admits invariant density $\pi_\eta(\theta)\propto \exp(\eta\ell(\theta))$.  Moreover, $\pi_\eta(\theta)$ concentrates on the set of global maximisers of $\ell$ as $\eta\to\infty$ \citep{hwang1980laplace}.

\section{THE ALGORITHM}\label{sec:algorithm}

In practice, simulating Langevin dynamics given in \eqref{eq:langevin_dynamics_for_optimisation} requires repeated evaluations of $\nabla_\theta \ell(\theta)$; we therefore turn to how this gradient can be estimated in the marginal likelihood setting.
We note that, under standard regularity conditions allowing differentiation under the integral sign, by using Fisher's identity \citep[Proposition~D.4]{douc2014nonlinear} for $\nabla_\theta \log p_\theta(y)$, we have
\begin{align}\label{eq:true-gradient}
\nabla_\theta \ell(\theta) &= \mathbb{E}_{p_{\text{data}}} \mathbb{E}_{p_\theta(\cdot | Y)}\left[ \nabla \log p_\theta(\cdot, Y) \right].
\end{align}
Replacing $p_{\text{data}}$ with the empirical distribution ${p}_{\text{data}}^M$, we can estimate the gradient in \eqref{eq:true-gradient} with
\begin{align*}
\nabla_\theta \ell_M(\theta) = \frac{1}{M} \sum_{m=1}^M \mathbb{E}_{p_\theta(\cdot | y_m)}\left[ \nabla \log p_\theta(\cdot, y_m) \right].
\end{align*}
This gradient remains intractable as $p_\theta(\cdot | y_m)$ is a posterior distribution which is typically not available in closed form. A typical workaround is to use \gls*{mcmc} to sample from $p_\theta(\cdot | y_m)$ \citep{pang2020ebmprior,debortoli2021soul}. We will instead follow \citet{akyildiz2023ipla}, \citet{kuntz2023} and use a particle approach where a set of $N$ particles will target the posterior $p_\theta(\cdot | y_m)$ for each data point $y_m$. We denote these particles as $X_t^{m,n}$, where $m \in \{1, \ldots, M\} $ indexes the data point and $n \in \{1, \dots, N\}$ indexes the particle.
\subsection{An Interacting Particle Algorithm}
For each fixed \(y\in\mathbb{R}^{d_y}\), define \(\phi_y:\mathbb{R}^{d_\theta}\times\mathbb{R}^{d_x}\to\mathbb{R}\) by \(\phi_y(\theta,x)=-\log p_\theta(x,y)\). For each data point $y_m$, we evolve $N$ particles, denoted $X_t^{m,n}$ for $n = 1, \ldots, N$, according to Langevin dynamics targeting the posterior $p_\theta(\cdot | y_m)$. Simultaneously, we use the particles to estimate the gradient $\nabla_\theta \ell_M(\theta)$ and evolve $\theta$ according to Langevin dynamics targeting the maximisers of $\ell_M(\theta)$. This can be ensured by setting the noise scaling in \eqref{eq:langevin_dynamics_for_optimisation} to be $\eta = {MN}$. With this choice,
as evident in our proof the associated joint Langevin system has a \(\theta\)-marginal proportional to \(\exp(MN\ell_M(\theta))\) which concentrates around the maximisers of $\ell_M(\theta)$ with increasing number of particles $N$.

Let us denote the joint empirical measure of the particles and data points as
$\mathsf{p}_t^{MN} = ({1}/{MN}) \sum_{m=1}^M \sum_{n=1}^N \delta_{(y_m, X_t^{m,n})}$. Using this, we propose to use the following SDE system:
\begin{align}
&\mathrm{d} \theta_t = - \E_{\mathsf{p}_t^{MN}}\left[\nabla_{\theta} \phi_{Y}(\theta_t, X) \right] \mathrm{d}t + \sqrt{\frac{2}{MN}} \mathrm{d}B_t, \label{eq:IPLD_1}\\
&\mathrm{d} X_t^{m,n} = - \nabla_{x} \phi_{y_m}(\theta_t, X_t^{m,n}) \mathrm{d}t + \sqrt{2} \mathrm{d}B_t^{m,n}, \label{eq:IPLD_2}
\end{align}
where $(B_t)_{t\geq 0}$ is a $d_\theta$-dimensional Brownian motion and $(B_t^{m,n})_{t\geq 0}$ for $m \in \{1, \ldots ,M\}$, $n \in \{1, \ldots N\}$ denote a family of $d_x$-dimensional independent standard Brownian motions. This is a generalisation of the interacting particle system introduced in \citet{akyildiz2023ipla} to the $M$ data point setting. One can observe that  Eq.~\eqref{eq:IPLD_1} aligns with the global optimisation setting given in Eq.~\eqref{eq:langevin_dynamics_for_optimisation} with the choice of $\eta = MN$.

In order to construct an implementable algorithm, we discretise the SDEs \eqref{eq:IPLD_1}--\eqref{eq:IPLD_2} using an Euler-Maruyama discretisation. Given a (possibly random) initial value $\theta_0$ and particles $X_0^{1:M,1:N}$, a step size $h>0$, the discretised system for $k \geq 0$ is given by
\begin{align}
&\theta_{k+1} = \theta_k - h \E_{\mathsf{p}_k^{MN}}\left[\nabla_{\theta} \phi_Y(\theta_k, X) \right] + \sqrt{\frac{2h}{MN}} W_k, \label{eq:ipla-algo-1} \\
&X_{k+1}^{m,n} = X_k^{m,n} - h \nabla_{x} \phi_{y_m}(\theta_k, X_k^{m,n}) + \sqrt{2h} W_k^{m,n}, \label{eq:ipla-algo-2}
\end{align}
where $\mathsf{p}_k^{MN} = ({1}/{MN}) \sum_{m=1}^M \sum_{n=1}^N \delta_{(y_m, X_k^{m,n})}$ and $(W_k)_{k\geq 0}$ and $(W_k^{m,n})_{k\geq 0}$ for all $m \in \{1, \ldots ,M\}$, $n \in \{1, \ldots N\}$ denote independent standard Gaussian random variables of appropriate dimension\footnote{With a slight abuse of notation, we use the same notation for the continuous time processes $(\theta_t, X_t^{m,n})$ and their discretisation $(\theta_k, X_k^{m,n})$.}. Specifically, the drift coefficient in \eqref{eq:ipla-algo-1} can be explicitly written as
\begin{align*}
\E_{\mathsf{p}_k^{MN}}\left[\nabla_{\theta} \phi_Y(\theta_k, X) \right] = \frac{1}{MN} \sum_{m=1}^M\sum_{n=1}^N \nabla_{\theta} \phi_{y_m}(\theta_k, X_k^{m,n}).
\end{align*}
To connect this to the estimation task for our model given in \eqref{eq:latent-ebm-model}, let us define the function $V_{\beta}(x, y) = - \log p_\beta(y | x)$. Given the energy function $U_\alpha:\mathbb{R}^{d_x} \to \mathbb{R}$, the gradient with respect to $\theta = (\alpha, \beta)$ is made explicit to clarify the implementation (for the derivation, see Appendix~\ref{app:gradient-derivations}). We can thus rewrite the system in Eqs.~\eqref{eq:ipla-algo-1}--\eqref{eq:ipla-algo-2} as
\begin{align}
    \alpha_{k+1} &= \alpha_k - \frac{h}{MN} \sum_{n=1}^N\sum_{m=1}^M  \nabla_\alpha U_{\alpha_k}(X_k^{m,n}) \nonumber  \\
    &  +  h\mathbb{E}_{p_{\alpha_k}(x)}[\nabla_\alpha U_{\alpha_k}(x)] + \sqrt{\frac{2h}{MN}} W^{\alpha}_k,\label{eq:alpha-update}\\
    \beta_{k+1} & = \beta_k -\frac{h}{MN} \sum_{n=1}^N \sum_{m=1}^M \nabla_\beta  V_{\beta_k}(X_k^{m,n}, y_m) \nonumber\\
    &+ \sqrt{\frac{2h}{MN}} W^{\beta}_k,\label{eq:beta-update} \\
    X_{k+1}^{m, n}  &= X_k^{m,n} -h\nabla_{x} U_{\alpha_k}(X_{k}^{m, n}) \nonumber \\
     &- h\nabla_{x} V_{\beta_k}(X_{k}^{m, n}, y_m) + \sqrt{2h}W_k^{m,n} \label{eq:x-update}
\end{align} 
for all $k \in \{0, \ldots, K-1\}$, where $(W^{\alpha}_k)_{k \ge 0}$, $(W^{\beta}_k)_{k \ge 0}$, and $(W_k^{m,n})_{k \ge 0}$ denote independent standard Gaussian random variables of appropriate dimension.  Note that the update for $\alpha$ in \eqref{eq:alpha-update} involves an expectation with respect to the prior, i.e. $\mathbb{E}_{p_{\alpha_k}(x)}[\nabla_\alpha U_{\alpha_k}(x)]$, which originates from the normalising constant $Z_\alpha$ (see Appendix~\ref{app:gradient-derivations} for details). As this expectation is generally intractable, it must also be approximated.

To approximate the expectation in Eq.~\eqref{eq:alpha-update}, we resort to a short-run \gls*{mcmc}-based approach, as typically done in training of \glspl*{ebm} \citep{song2021train} and \glspl*{lebm} \citep{pang2020ebmprior}. For each data point, we generate approximate prior samples using a short-run \gls*{ula} chain targeting the prior distribution $p_{\alpha}$. 
Specifically, for each data point $y_m$ and a fixed parameter value $\alpha_k$, we initialise the chain at $\hat{X}_{k,0}^m \sim p_0$, where $p_0 = \mathcal{N}(0,I_{d_x})$, and evolve it for a fixed number of steps $J$ with a step-size $\gamma > 0$: 
\begin{align}\label{eq:prior-langevin-dynamics}
    \hat{X}_{k,j+1}^m &= \hat{X}_{k,j}^m - \gamma \nabla_x U_{\alpha_k}(\hat{X}_{k,j}^m) + \sqrt{2\gamma}W_{k,j}^{m},
\end{align}
where $(W_{k,j}^m)_{j,k\geq 0}$ for $m \in \{1, \ldots, M\}$ denote independent standard $d_x$-dimensional Gaussian random variables. Then for every $k \geq 0$ and for each data point $y_m$, we run an \gls*{mcmc} chain for $J$ steps and we obtain the following system of equations
\begin{align}
    \tilde{\alpha}_{k+1} &= \tilde{\alpha}_k - \frac{h}{MN} \sum_{n=1}^N\sum_{m=1}^M  
    \nabla_\alpha U_{\tilde{\alpha}_k}(\tilde{X}_k^{m,n}) \nonumber  \\
    &\quad +  \frac{h}{M}\sum_{m=1}^M \nabla_\alpha U_{\tilde{\alpha}_k}(\hat{X}^{m}_{k,J}) 
    + \sqrt{\tfrac{2h}{MN}}\, \tilde{W}^{\alpha}_k,\label{eq:alpha-update-inexact}\\
    \tilde{\beta}_{k+1} & = \tilde{\beta}_k -\frac{h}{MN} \sum_{n=1}^N \sum_{m=1}^M 
    \nabla_\beta V_{\tilde{\beta}_k}(\tilde{X}_k^{m,n}, y_m) \nonumber\\
    &\quad+ \sqrt{\tfrac{2h}{MN}}\, \tilde{W}^{\beta}_k,\label{eq:beta-update-inexact} \\
    \tilde{X}_{k+1}^{m, n}  &= \tilde{X}_k^{m,n} -h\nabla_{x} U_{\tilde{\alpha}_k}(\tilde{X}_{k}^{m, n}) \nonumber \\
    &\quad - h\nabla_{x} V_{\tilde{\beta}_k}(\tilde{X}_{k}^{m, n}, y_m) 
    + \sqrt{2h}\,\tilde{W}_k^{m,n}, \label{eq:x-update-inexact}
\end{align}
where $(\tilde{W}^{\alpha}_k)_{k \ge 0}$, $(\tilde{W}^{\beta}_k)_{k \ge 0}$, and $(\tilde{W}_k^{m,n})_{k \ge 0}$ denote independent standard Gaussian random variables of appropriate dimension. The complete algorithm is summarised in Algorithm~\ref{algo:simple}.
\begin{algorithm}[t]
    \caption{Energy-Based Interacting Particle Langevin Algorithm (Full version)}\label{algo:simple}
    \begin{algorithmic}[1]
    \Require \hfill \vphantom{x}
\begin{tabular}{@{}l p{0.63\linewidth}}
    $\{y_m\}_{m=1}^M, M$ &  {\footnotesize\textcolor{mypink}{\texttt{\# Dataset and its size}}} 
    \\
  $K$  & {\footnotesize\textcolor{mypink}{\texttt{\# Number of iterations}}}
  \\
  $N$    & {\footnotesize\textcolor{mypink}{\texttt{\# Particles per data point}}} \\
  $J$                 & {\footnotesize\textcolor{mypink}{\texttt{\# Prior sampling steps}}} \\
  $h > 0$             & {\footnotesize \textcolor{mypink}{\texttt{\# Step-size for discretisation}}} \\
    $\gamma > 0$        & {\footnotesize\textcolor{mypink}{\texttt{\# Step-size for prior sampling}}} \\
    $\tilde{\theta}_0 = (\tilde{\alpha}_0, \tilde{\beta}_0)$ & {\footnotesize\textcolor{mypink}{\texttt{\# Initial parameters values}}} \\
    $\{\tilde{X}_0^{m,n}\}_{m,n=1}^{M,N}$ & {\footnotesize\textcolor{mypink}{\texttt{\# Initial particles}}}
\end{tabular}
\For{$k = 0$ to $K-1$}
\State Sample $\hat{X}_{k,J}^m$ using $J$ steps of \eqref{eq:prior-langevin-dynamics} with $\tilde{\alpha}_k$ for each $m \in \{1, \ldots, M\}$.

\State Update $\tilde{\alpha}_{k+1}$ according to Eq.~\eqref{eq:alpha-update-inexact} 
\State Update $\tilde{\beta}_{k+1}$ according to Eq.~\eqref{eq:beta-update-inexact}
\For{$(m,n) \in \{1, \ldots, M\} \times \{1, \ldots, N\}$} 
\State Update $\tilde{X}_{k+1}^{m,n}$ according to Eq.~\eqref{eq:x-update-inexact}
\EndFor
\EndFor
\end{algorithmic}
\end{algorithm}
\subsection{Practical Implementation Details}
In practice, we use several standard techniques to ensure stable and efficient training (see Appendix~\ref{app:practical-algorithm} for details). To optimise the parameters $(\alpha,\beta)$, we use the Adam optimiser \citep{kingma2017adammethodstochasticoptimization}, which uses adaptive step sizes and momentum and has become a standard choice for training deep models due to its robustness across different scales of parameters. To handle large datasets efficiently, we rely on mini-batching: instead of using the full dataset at each iteration, we approximate the gradients using randomly sampled subsets of data \citep{robbins1951stochastic}. This reduces computational cost and gives an unbiased estimator of the corresponding finite-sum gradient. Finally, since mini-batching implies that the decoder's parameters are updated per batch rather than per full dataset pass, we introduce a suitable noise addition scheme for the posterior particles to approximately match the full-data Euler-Maruyama discretisation over one epoch. The complete algorithm and further details are provided in Appendix~\ref{app:practical-algorithm}.

\section{NONASYMPTOTIC ANALYSIS}\label{sec:nonasymptotic}
In this section, we study convergence of \gls*{iplebm} to the global maximisers of the marginal log-likelihood under strong convexity and $L$-smoothness assumptions. We focus on how convergence depends on both the number of particles $N$ and the number of data points $M$. We first consider the case where the prior expectation $\mathbb{E}_{p_{\alpha_k}}[\nabla_\alpha U_{\alpha_k}(x)]$ appearing in Eq.~\eqref{eq:alpha-update} is available exactly, and then handle the inexact setting where it is replaced by a biased MCMC estimate.
\subsection{Assumptions}
We first assume the negative joint log-likelihood is strongly convex and $L$-smooth for every data point.
\begin{assumption}\label{ass:strong-convexity} (Strong convexity) Let $v = (\theta, x)$ and $v' = (\theta', x')$. We assume that there exists $\mu > 0$ such that
    \begin{equation*}
        \langle\nabla \phi_y(v) - \nabla \phi_y (v'), v-v' \rangle \ge \mu \|v - v' \|^2
    \end{equation*}
for all $y \in \mathbb{R}^{d_y}$ and $v, v' \in \mathbb{R}^{d_\theta} \times \mathbb{R}^{d_x}$.
\end{assumption}
\begin{remark}[Log-concavity of the marginal likelihood]\label{remark:strong-convexity} Note that $\phi_{y_m}(\theta, x) = - \log p_\theta(x, y_m)$ is the negative joint log-likelihood, thus \Cref{ass:strong-convexity} is a joint log-concavity assumption that holds for $p_\theta(x, y_m)$ for every $m$.Since $\phi_{y_m}(\theta,\cdot)$ is $\mu$-strongly convex, $p_\theta(y_m) < \infty$ for all $\theta$, and by Pr\'ekopa's theorem it can be shown that $p_\theta(y_m)$ is $\mu$-strongly log-concave in $\theta$.
This then implies that $\ell_M(\theta)$ is also $\mu$-strongly concave and hence has a unique maximiser.
\end{remark}

\begin{assumption}\label{ass:L-smooth} ($L$-smoothness) Let $v = (\theta, x)$. We assume that there exists $L > 0$ such that
    \begin{equation*}
        \|\nabla \phi_y(v) - \nabla \phi_y (v') \| \le L\|v - v' \|
    \end{equation*}
    for all $y \in \mathbb{R}^{d_y}$ and $v, v' \in \mathbb{R}^{d_\theta} \times \mathbb{R}^{d_x}$
\end{assumption}

\subsection{Convergence with exact gradient}
We first analyse the simplified setting in Eqs.~\eqref{eq:ipla-algo-1}--\eqref{eq:ipla-algo-2}. In other words, we assume that, equivalently, in Eqs.~\eqref{eq:alpha-update}--\eqref{eq:x-update}), the expectation w.r.t. $p_{\alpha_k}$ is known. This allows us to isolate the core convergence behaviour of the particle-based updates as can be seen below.
\begin{theorem}\label{thm:exact-gradient}
Suppose \Cref{ass:strong-convexity}, \Cref{ass:L-smooth} hold. Let $\theta_k$ be the parameter marginal generated by iterates \eqref{eq:ipla-algo-1}--\eqref{eq:ipla-algo-2} (equivalently, \eqref{eq:alpha-update}--\eqref{eq:x-update}), then given a step size $0 < h \leq 2 / (\mu + L)$, the following holds
\begin{align*}
&\mathbb{E}\left[\|\theta_k - \theta_\star\|^2\right]^{1/2} \leq (1-\mu h)^k C_0 + C_1 h^{1/2} + \frac{C_2}{\sqrt{MN}},
\end{align*}
where $C_0$ is an explicit constant which depends on the initial law of the system and
\begin{align*}
C_1 &= 1.65 \frac{L}{\mu} \sqrt{\frac{d_\theta + M N d_x}{MN}}, \quad C_2 = \sqrt{\frac{d_\theta}{\mu}},
\end{align*}
and $\theta_\star$ denotes the unique maximiser of $\ell_M$.
\end{theorem}
\begin{proof}
A full proof can be found in Appendix~\ref{sec_app:proof-exact-gradient}.
\end{proof}
\begin{remark}[Convergence rate]\label{remark:thm-exact-gradient}
    Theorem~\ref{thm:exact-gradient} implies that, for sufficiently large $N$ and $k$ and sufficiently small $h$, the iterates of Eqs.~\eqref{eq:alpha-update}--\eqref{eq:x-update} can be made arbitrarily close to $\theta_\star$. More precisely for a fixed number of data points $M$ and a target accuracy $\varepsilon>0$  choosing $N = \mathcal{O}(\varepsilon^{-2}M^{-1} d_\theta)$ makes the last term of the bound in Theorem~\ref{thm:exact-gradient} of order $\mathcal{O}(\varepsilon)$. Next, setting $h = \mathcal{O}(\varepsilon^2 d_x^{-1})$ makes the middle term  of order $\mathcal{O}(\varepsilon)$. Finally setting $k \ge \widetilde{\mathcal{O}}(d_x \varepsilon^{-2})$ ensures that the first term is $\mathcal{O}(\varepsilon)$. With these choices, we obtain $\mathbb{E}[\|\theta_k-\theta_\star\|^2]^{1/2} \leq \mathcal{O}(\varepsilon)$. Note that since $N = \mathcal{O}(M^{-1}\varepsilon^{-2})$, for large datasets the bound can be made small with small $N$.
\end{remark}

Below, we provide a sketch of the proof. We start by noting that $\mathbb{E}[\|\theta_k-\theta_{\star}\|^2]^{1 / 2}=W_2(\mathcal{L}(\theta_k), \delta_{\theta_{\star}})$ since $\delta_{\theta_{\star}}$ is a Dirac measure \citep[Section 1.4]{santambrogio2015optimal}. Then via an application of the triangle inequality, we obtain:
\begin{align*}
\mathbb{E}\left[\left\|\theta_k-\theta_{\star}\right\|^2\right]^{1 / 2} & =W_2\left(\mathcal{L}\left(\theta_k\right), \delta_{\theta_{\star}}\right) \\
& \leq \underbrace{W_2\left(\pi_{\Theta}, \delta_{\theta_{\star}}\right)}_{\text {concentration }}+\underbrace{W_2\left(\mathcal{L}\left(\theta_k\right), \pi_{\Theta}\right)}_{\text {convergence }}.
\end{align*}
where $\pi_\Theta$ denotes the $\theta$ marginal of the invariant measure of the analysed system of SDEs. We establish concentration using \citet[Lemma~A.8]{altschuler2024faster} and convergence by adapting \citet[Theorem~1]{dalalyan2019user} for our system.

\begin{remark}[Non-convex setting]
It is important to note that the $\theta$-marginal of the stationary measure of the proposed SDE system in \eqref{eq:IPLD_1}-\eqref{eq:IPLD_2} concentrates on the maximisers of the likelihood $\ell_M(\theta)$ with increasing $N$ under Laplace-type regularity conditions~\citep{hwang1980laplace} on $\ell_M$ without the need for the strong-convexity assumption \Cref{ass:strong-convexity}. This can be readily observed in the proof of Theorem \ref{thm:exact-gradient} (see Appendix \ref{sec_app:proof-exact-gradient}), where the stationary measure of the joint system is shown to have the desired marginal properties. Thus, while \Cref{ass:strong-convexity} is used to establish the non-asymptotic convergence rates presented in this section, the algorithm is inherently designed to target the \gls*{mmle} solution in more general landscapes.
\end{remark}
\subsection{Convergence with inexact gradient}
Since the expectation with respect to the prior $p_\alpha$ cannot be computed exactly, 
the gradient update in Eq.~\eqref{eq:alpha-update} is not directly available. 
To address this, we approximate the expectation using a \gls*{ula} chain, which yields 
a biased estimator of  $\mathbb{E}_{p_{\alpha_k}(x)}\big[\nabla_\alpha U_{\alpha_k}(x)\big]$ at each iteration. Following \cite{dalalyan2019user}, we impose the assumptions below to control the bias introduced by this approximation and the variance of the deviation from the true quantity.

Let $g_k \coloneqq ({1}/{M})\sum_{m=1}^M\nabla_\alpha U_{\tilde{\alpha}_k}(\hat{X}^{m}_{k,J})$ be the estimate of $\mathbb{E}_{p_{\tilde{\alpha}_k}(x)}[\nabla_\alpha U_{\tilde{\alpha}_k}(x)]$ obtained using the \gls*{ula} samples as defined in Eq.~\eqref{eq:alpha-update-inexact}. Let $\zeta_k \coloneqq g_k - \mathbb{E}_{p_{\tilde{\alpha}_k}(x)}[\nabla_\alpha U_{\tilde{\alpha}_k}(x)]$ denote the difference between the estimator and the true expectation value for all $k\geq0$. Finally, let $\mathcal{F}_k$ be the $\sigma$-algebra generated by the algorithm (Eqs.~\eqref{eq:alpha-update-inexact}--\eqref{eq:x-update-inexact}) up to iteration $k$.
\begin{assumption}\label{ass:bound-on-bias}
    There exist $\delta >  0$ and $\sigma > 0$ such that
    \begin{align*}
        \mathbb{E}\left[\left\|\mathbb{E}\left[\zeta_k | \mathcal{F}_k \right]\right\|^2\right] &\le \delta^2, \quad
        \mathbb{E}\left[\left\|\zeta_k - \mathbb{E}\left[\zeta_k | \mathcal{F}_k \right]\right\|^2\right] &\le \sigma^2
    \end{align*}
   for $k \geq 0$, where $\tilde{W}^\alpha_{k+1}$, $\tilde{W}^\beta_{k+1}$, $\tilde{W}_{k+1}^{m,n}$ for all $m \in \{1, \ldots ,M\}$, $n \in \{1, \ldots N\}$ in Eqs.~\eqref{eq:alpha-update-inexact}--\eqref{eq:x-update-inexact} are independent of $(\zeta_0, \ldots \zeta_k)$.
\end{assumption}

\begin{remark}
    Intuitively, the bias $\zeta_k$ arises from a finite run of \gls*{ula}. Here $\delta := \delta(J)$ decreases with the number of \gls*{ula} iterations $J$, and both $\delta$ and $\sigma$ implicitly depend on $d_\alpha$ as the gradient error mainly arises in the $\alpha$ component. While not directly applicable to our setting - which involves inexact gradients and averaging across parallel chains - the decomposition of \citet{durmus2017} under geometric ergodicity offers useful intuition on how $\delta$ depends on $J$: \gls*{ula} error splits into an irreducible $O(\gamma)$ discretisation bias and a decaying initialisation term which decays geometrically in $J$. This motivates choosing $\gamma$ small enough to control the bias from inexactly targeting $p_\alpha$. Hence, if the \gls*{ula} chain is stable, geometrically ergodic and $J$ is large enough while $\gamma$ is small enough, \cref{ass:bound-on-bias} can hold in practice.
\end{remark}
 In this setting we can get a similar bound as in Theorem \ref{thm:exact-gradient}  with additional terms arising from the bias in the expectation estimate.
\begin{theorem}\label{thm:inexact-gradient}
Suppose \Cref{ass:strong-convexity}, \Cref{ass:L-smooth}, \Cref{ass:bound-on-bias} hold. Let $\tilde{\theta}_k = (\tilde{\alpha}_k, \tilde{\beta}_k)$ be the parameter marginal generated by iterates \eqref{eq:alpha-update-inexact}--\eqref{eq:x-update-inexact}, then given a step size $ 0 < h \leq {2}/({\mu + L})$, the following holds
\begin{align*}
\mathbb{E}\left[\|\tilde{\theta}_k - \theta_\star\|^2\right]^{1/2} \leq (1-\mu h)^k \tilde{C}_0 &+ \tilde{C}_1 h^{1/2} \\
&+\frac{\tilde{C}_2}{\sqrt{MN}} + \tilde{C}_3,
\end{align*}
where $\tilde{C}_0$ is a fully explicit constant that only depends on the initial law of the system, $\tilde{C}_1$ is an explicit constant that depends on $M, N, \sigma, \mu,  d_\theta$ and $d_x$ and stabilises as $M, N \to \infty$,  $\tilde{C}_2$ is an explicit constant that depends on $\mu$ and $d_\theta$, and
$\tilde{C}_3$ is an explicit constant that depends on $\mu$ and $\delta$.
\end{theorem}
\begin{proof}
See Appendix~\ref{sec_app:proof-inexact-gradient} for the proof with full expressions of constants.
\end{proof}

The proof follows similar steps to that of Theorem~\ref{thm:exact-gradient}, 
adapted to our setting by applying Theorem~4 from \cite{dalalyan2019user}. Crucially, 
the resulting bound includes an irreducible bias term that does not vanish as 
$M, N \rightarrow \infty$. As discussed above, this term originates directly from 
the irreducible bias $\delta(J)$.

\section{RELATED WORKS}
\begin{figure*}[htbp]
    \centering
    \begin{minipage}[c]{0.78\textwidth}
        \includegraphics[width=\linewidth]{figures/toy_exp/particle_and_energy_combined.pdf}
    \end{minipage}
    \hfill
    \begin{minipage}[c]{0.2\textwidth}
        \centering
        \includegraphics[width=\linewidth]{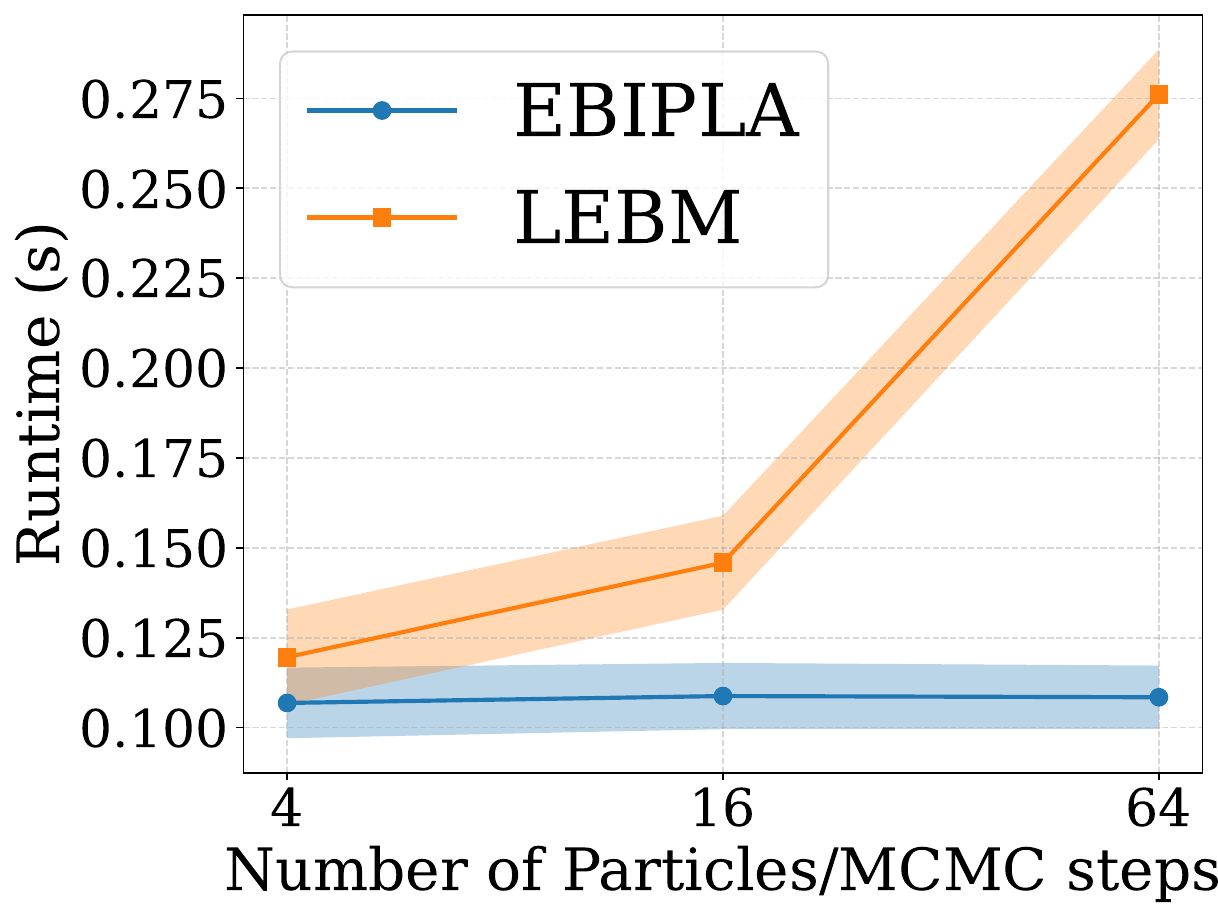}
    \end{minipage}
    \vspace{0pt} 
    \caption{ \textbf{Left:} Samples generated with \gls*{iplebm} and \gls*{lebm} trained on three synthetic datasets for increasing numbers of posterior particles/MCMC steps together with the corresponding exponentiated energy landscapes. Sample quality and energy accuracy improve with larger particle counts/iterations. The two methods achieve similar performance in terms of sample quality, however \gls*{iplebm} learns a more realistic energy landscape \textbf{Right:} Comparison of runtime for increasing number of posterior particles (\gls*{iplebm}) or MCMC steps (\gls*{lebm}). \gls*{iplebm} consistently achieves comparative qualitative performance while being substantially faster: its runtime scales sub-linearly with the number of particles, remaining below \gls*{lebm} even at the largest particle count.}
    \label{fig:combined_results}
\end{figure*}
\paragraph{Training EBMs.}

Efficient training of \glspl*{ebm} remains a fundamental challenge in generative modelling. As an approximation to \gls*{mle}, \citet{hinton2002training} introduce Contrastive Divergence (CD) and \citep{tieleman2008} follow up with Persistent CD (PCD) which use short-run chains, though these introduce uncontrolled bias \citep{nijkamp2020anatomy}. Recent alternatives include non-equilibrium thermodynamic approaches \citep{carbone2024efficient,cuin2026efficient} and the Energy Discrepancy loss \citep{schroeder2023energy, schroeder2024energybased}, which bypasses \gls*{mcmc} by optimising a different objective. \gls*{iplebm} differentiates itself by providing a diffusion-based alternative in the latent setting of \citet{pang2020ebmprior}. Based on a novel SDE framework, our Euler-Maruyama discretisation replaces sequential posterior \gls*{mcmc} with simultaneous particle updates. While this resembles a one-step form of PCD, it arises from a distinct theoretical foundation. To our knowledge, the only other work providing a diffusion limit for \gls*{ebm} training is \citet{oliva2025uniformintime}, though they focus exclusively on the non-latent setting.

\paragraph{EBMs in the Latent Space.} 
Recently, several works studying latent \glspl*{ebm} have appeared. Even though they focus on the study of latent \glspl*{ebm} they are quite different in spirit to our method. \cite{yu2023diffusionamortized} modify the \gls*{lebm} setting and train an additional diffusion model to facilitate training. Although they obtain superior results, their method is substantially more expensive than the classic \cite{pang2020ebmprior}. \cite{xiao2021vaebm} train an \gls*{ebm} in the latent space of a pre-trained VAE to motivate the use of an \gls*{ebm} as a prior, as they show it enhances performance of the VAE. \cite{xiao2022adaptive} suggest using a multi-stage noise contrastive estimation instead of improving the classical likelihood training as we do in our work. Finally, \cite{2025cuimultimodallatent} propose a variational learning scheme for the generator posterior, which is different from our particle based approach, in order to extend the training method to a multimodal setting.

\paragraph{Interacting Particle Algorithms for Learning.}
\citet{kuntz2023, caprio2024} introduce Particle Gradient Descent (PGD), a diffusion-based particle algorithm for \gls*{mmle} in latent variable models that jointly optimises model parameters and samples from the posterior distribution of the latent variables. Building on this framework, \citet{akyildiz2023ipla} incorporate additive noise in the parameter updates and establish associated non-asymptotic convergence guarantees which we build upon and extend to multiple data-points and inexact gradient setting. Since then, particle-based learning methods have been taking off in various directions. Momentum-enriched variants of \gls*{ipla} \citep{lim2024momentum, oliva2024} and extensions to non-smooth settings following \cite{encinar2025proximal} offer natural extensions to \gls*{iplebm}. From a geometric viewpoint, MMLE can be seen as a discretisation of a gradient flow in Wasserstein-2 space \citep{kuntz2023} , which suggests potential generalisations based on alternative geometries, as explored by \cite{sharrock2024tuning} for Stein variational gradient descent, which could be another potential extension of our method. We also mention recent \gls*{smc}-based approaches for learning \citep{carbone2024efficient,crucinio2025mirror,cuin2025learning,cuin2026efficient}, which can be used to develop generalisations of our work where particles have weights.

\section{EXPERIMENTS}\label{sec:experiments}
\begin{table*}
    \centering
    \scriptsize
    \begin{tabular}{l r r r r r r}
                \toprule
                \textbf{Model} & \multicolumn{2}{c}{\textbf{SVHN}} & \multicolumn{2}{c}{\textbf{CIFAR-10}} & \multicolumn{2}{c}{\textbf{CelebA64}} \\
                \cmidrule(lr){2-3} \cmidrule(lr){4-5} \cmidrule(lr){6-7}
                & {MSE}$\downarrow$ & {FID}$\downarrow$ & {MSE}$\downarrow$ & {FID}$\downarrow$ & {MSE}$\downarrow$ & {FID}$\downarrow$ \\
                \midrule
                VAE \citep{kingma2013auto}             & $0.019$ & $46.78$ & $0.057$ & $106.37$ & $0.021$ & $65.75$ \\
                2s-VAE \citep{dai2019diagnosing}       & $0.019$ & $42.81$ & $0.056$ & $72.90$  & $0.021$ & $44.40$ \\
                RAE \citep{ghosh2019variational}       & $0.014$ & $40.02$ & $0.027$ & $74.16$  & $0.018$ & $40.95$ \\
                SRI ($L=5$) \citep{nijkamp2020learning}  & $0.011$ & $35.32$ & --    & --     & $0.015$ & $47.95$ \\
                \gls*{lebm} \citep{pang2020ebmprior}       & $0.008$ & $29.44$ & $\underline{0.020}$ & $\mathbf{70.15}$ & $0.013$ & $37.87$ \\
                SM-LEBM \citep{schroeder2023energy} & $0.010$ & $34.44$ & $0.026$ & $77.82$ & $0.014$ & $41.21$ \\
                ED-LEBM \citep{schroeder2023energy} & $\underline{0.006}$ & $\underline{28.10}$ & $0.023$ & $\underline{73.58}$ & $\mathbf{0.009}$ & $\underline{36.73}$ \\
                \midrule 
                \gls*{iplebm} (ours)    & $\mathbf{0.004_{\pm {4e-5}}}$  &  $\mathbf{27.54_{\pm 0.42}}$  & $\mathbf{0.017_{\pm 5e-4}}$  & $75.13_{\pm 0.74}$  & $\underline{0.013_{\pm 9e-5}}$ & $\mathbf{35.72_{\pm 0.46}}$  \\ 
                \bottomrule
            \end{tabular}
            \caption{Comparison of MSE($\downarrow$) and FID($\downarrow$) on the SVHN, CIFAR-10, and CelebA64 datasets. We use bold font to indicate the best performance and underline to indicate the second best performance. We report the mean and standard error of the metrics averaged over 5 runs.}
            \label{table:image-reconstruction-generation-simple}
\end{table*}

\begin{figure*}[htbp]
    \centering
    \includegraphics[width=0.6\linewidth]{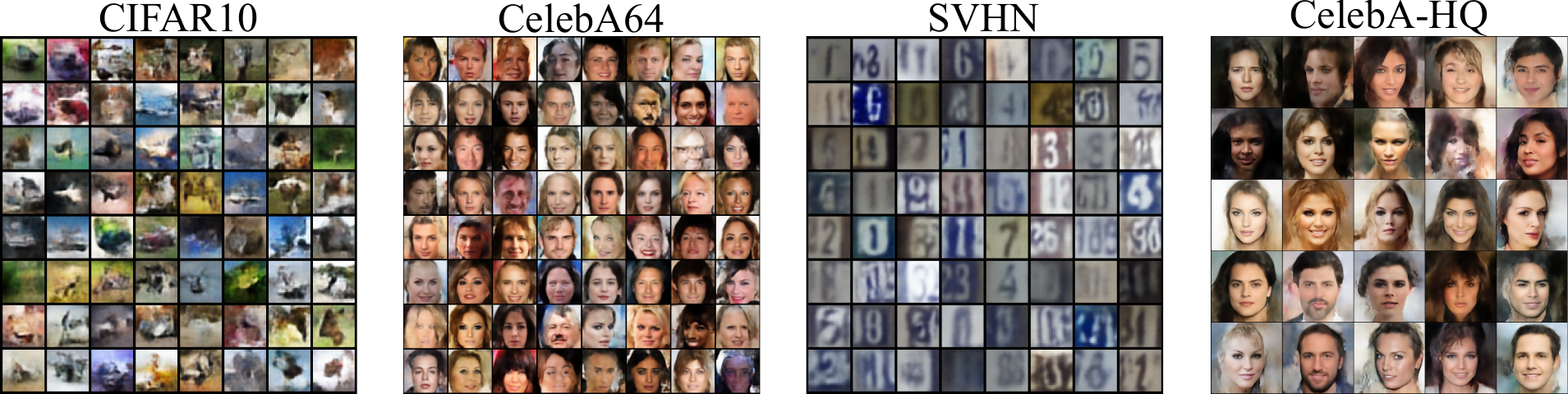}
    \caption{Samples generated with \gls*{iplebm} trained on CIFAR-10, CelebA64, SVHN, and CelebA-HQ.}
    \label{fig:generated_samples}
\end{figure*}
\begin{figure}[htbp]
    \centering
    \captionsetup[subfigure]{labelformat=simple,labelsep=space}
    \renewcommand\thesubfigure{Table~2(\alph{subfigure})}
    \makeatletter
    \renewcommand\p@subfigure{}
    \makeatother
    \begin{subfigure}[t]{0.48\linewidth}
        \centering
        \resizebox{\linewidth}{!}{
        \begin{tabular}{lc}
            \toprule
            \textbf{Model} & \textbf{FID}$\downarrow$ \\
            \midrule
            VAE        & $180.49$ \\
            ABP        & $160.21$ \\
            LEBM       & $133.07$ \\
            DAMC \citep{yu2023diffusionamortized} & $85.88$  \\
            \midrule 
            EBIPLA (ours) & $92.10$  \\
            \bottomrule
        \end{tabular}
        }
        \caption{Comparison of FID on the CelebA-HQ dataset. We do not report standard deviation due to computational constraints.}
    \end{subfigure}
    \hfill
    \begin{subfigure}[t]{0.48\linewidth}
        \centering
        \resizebox{\linewidth}{!}{
            \begin{tabular}{cc}
                \toprule
                \textbf{No. of Particles} & \textbf{FID}$\downarrow$ \\
                \midrule
                $N=1$  & $31.00_{\pm 0.81}$ \\
                $N=4$  & $27.83_{\pm 0.38}$ \\
                $N=16$  & $27.09_{\pm 0.31}$ \\
                $N=32$ & $26.74_{\pm 0.22}$ \\
                \bottomrule
            \end{tabular}
            }
        \caption{Effect of the number of posterior particles $(N)$ on the generation performance of \gls*{iplebm} trained on SVHN averaged over 5 runs.}
        \label{tab:svhn_ablation}
    \end{subfigure}
\end{figure}

To evaluate the generative performance of \gls*{iplebm} we train it on three synthetic datasets (Section \ref{sec:synthetic-datasets}) and on three image datasets
(Section \ref{sec:image-experiments}). On the synthetic task we benchmark against the closest baseline, \gls*{lebm} \citep{pang2020ebmprior}. Compared to the benchmark across both settings, \gls*{iplebm} attains slightly better generative performance while delivering substantially lower runtime at matched compute budgets. For the image datasets, we additionally compare to other \glspl*{lvm}.

\subsection{Synthetic datasets}\label{sec:synthetic-datasets}
We evaluate \gls*{iplebm} on three synthetic datasets: rotated Swiss roll, rotated Half-moons and rescaled Circle. More precisely, we generate a Swiss roll/Half-moons/Circle in a 2-dimensional latent space $\mathbb{R}^2$ and then map it to the ambient space through an orthogonal transformation $T : \mathbb{R}^2 \to \mathbb{R}^2$ (for Swiss roll and Half-moons) or linear transformation  $T : \mathbb{R}^2 \to \mathbb{R}^2$ for the Circle (for details see \Cref{app:synthetic-experiment-details}).

To contextualise our results, we consider the closest baseline, \gls*{lebm}, for comparison. To ensure a fair comparison, we match the computational budgets of both models so that they require the same number of gradient evaluations per training iteration. Specifically, the computational complexity for both \gls*{iplebm} and \gls*{lebm} is $\mathcal{O}(B(N+J))$, where $B$ is the batch size and $J$ denotes the number of prior iterations. To choose an appropriate value of $J$ we run an empirical study of the effect of $J$ on the performance of the algorithm (see Appendix \ref{app:additional-results}). We select $J =500$ as a practical trade-off between sample/energy quality and computational cost.  By setting $N \in \{4, 16, 64\}$ to represent the number of posterior particles for \gls*{iplebm} and the number of posterior MCMC steps for \gls*{lebm}, we ensure that the same number of gradient calls for the two methods are needed. We train both models for 200 epochs and report per training-step runtime for the Swiss roll dataset averaged over 20 runs (Figure \ref{fig:combined_results}).

\paragraph{Sample Quality.} In Figure \ref{fig:combined_results}, we can see that the sample quality and the accuracy of the learned energy landscape improves with increasing number of posterior particles/iterations. \gls*{iplebm} performs comparably in terms of sample quality to \gls*{lebm}. While \gls*{iplebm} learns the energy landscape more accurately, \gls*{lebm} results in distorted energy landscapes.

\paragraph{Computational Efficiency.} Figure \ref{fig:combined_results} shows that even though \gls*{iplebm} achieves comparable qualitative performance, empirically \gls*{iplebm} is significantly faster: its runtime grows sub-linearly with particle count, whereas \gls*{lebm}’s runtime increases  faster than linear with iteration count. This discrepancy stems from the underlying implementation: while \gls*{lebm} requires sequential posterior updates (typically implemented via a \texttt{for}-loop), \gls*{iplebm} updates all particles simultaneously. In terms of arithmetic gradient evaluations, both methods scale as $\mathcal{O}(B(N+J))$ under the matched-budget setup; however, because the $N$ posterior-particle updates in \gls*{iplebm} are vectorised, the observed wall-clock time scales as $\mathcal{O}(BJ)$ for \gls*{iplebm}.

\subsection{Image Modelling}\label{sec:image-experiments}

To validate the effectiveness of our method on high dimensional data, we train \gls*{iplebm} on three standard image datasets: CIFAR10 \citep{krizhevsky2009learning}, SVHN \citep{svhn2011}, and CelebA64 \citep{liu2015faceattributes}. Following \citet{pang2020ebmprior}, we use the same architecture for both the energy-based prior and the generator network. We train all models for $200$ epochs with a batch size of $128$; the number of particles is set to $N=10$ to balance performance and efficiency following \citet{kuntz2023}. We additionally test our method on high resolution CelebA-HQ $256\times 256$ datasets~\citep{liu2015faceattributes,karras2018progressive} using $N=5$ particles to demonstrate its scalability. The complete implementation details and training hyperparameters are provided in \Cref{app:image_details}. 

\paragraph{Generation and Reconstruction.} We evaluate the generative performance of our model quantitatively with the \gls*{FID} \citep{heusel2017gan} using $50,000$ samples generated by the energy-based prior and decoded back to the pixel space. Additionally, we measure the reconstruction quality in terms of mean-squared error (MSE) on the validation set. We obtain the reconstructions as the maximum a posteriori estimates of $p_{\theta}(x|y)$ (cf. \Cref{sec:algorithm}) following the implementation in \citet{kuntz2023}; we provide further details of the evaluation procedure in \Cref{app:image_details}. In \Cref{table:image-reconstruction-generation-simple}, we show that our \gls*{iplebm} achieves comparable performance among relevant baselines. {We focus on comparisons with latent \gls*{ebm} baselines, which allow us to isolate the impact of the introduced training method on the generative and reconstructive performance (for comparison with broader model selection see Section~\ref{app:additional-results}).} On CIFAR-10, our model is least competitive but remains comparable; we attribute this primarily to the simple Euler-Maruyama discretisation used in our implementation, rather than a fundamental limitation of the approach, noting that stronger baselines such as LEBM use substantially longer posterior chains (cf.~\Cref{tab:image_wallclock_1par}). We highlight that \gls*{iplebm} has shorter runtimes compared to LEBM \citep{pang2020ebmprior} across settings due to its efficient formulation (cf.~\Cref{tab:image_wallclock_1par}). {For the high resolution experiments, we additionally benchmark against~\cite{yu2023diffusionamortized}, which is a more complex model that uses an additional latent diffusion model to aid training and sampling of \gls*{lebm}; despite its simplicity, \gls*{iplebm} approaches the performance of this stronger baseline while substantially outperforming \gls*{lebm}.}
 Furthermore, we show on the SVHN dataset that the performance of \gls*{iplebm} improves as we increase the number of posterior particles used in~\ref{tab:svhn_ablation}. These empirical trends are consistent with the qualitative prediction of our theory: increasing the number of posterior particles reduces the particle approximation error. Although the image experiments do not necessarily satisfy the strong log-concavity and smoothness assumptions (\Cref{ass:strong-convexity}-\ref{ass:bound-on-bias}) used to obtain the non-asymptotic bounds in~\Cref{sec:nonasymptotic}, this behaviour is still expected since the underlying particle system is designed to concentrate around maximisers of the marginal likelihood beyond the analysed setting.
 
\begin{figure}[htbp]
    \centering
    \includegraphics[width=0.8\linewidth]{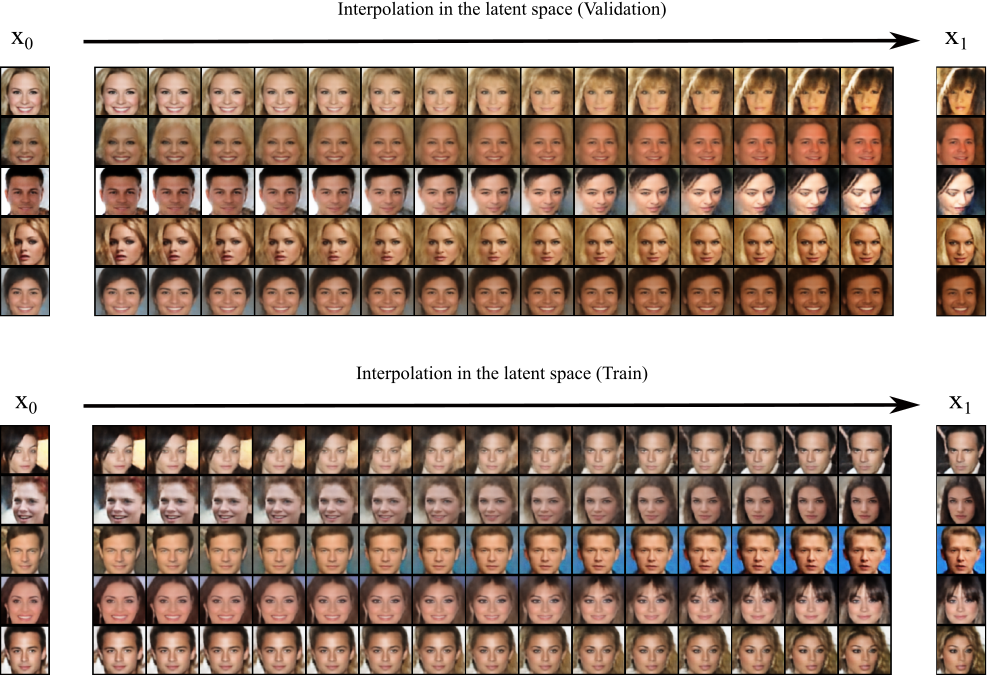}
    \caption{Latent space interpolation of \gls*{iplebm} trained on CelebA64 on the training (bottom) and validation set (top).}
    \label{fig:latent-space-interp}
\end{figure}

\paragraph{Latent Representations.} To qualitatively assess the learned representations, we visualise interpolations in the latent space. We linearly interpolate between the latent vectors $x_0$ and $x_1$ as $\texttt{lerp}(x_0, x_1, t)=tx_0+(1-t)x_1$; we then decode those interpolants back to the pixel space using the trained generator. For training images, we select index $m$ and extract one of the posterior particles randomly from the $N$ particles $X_{K}^{m,1:N}$ obtained from training (cf. \Cref{algo:simple}); for validation images, we use the MAP estimate as in our reconstruction experiments. \Cref{fig:latent-space-interp} shows the resulting interpolation trajectories, which suggests that \gls*{iplebm} has learned a smooth and semantically meaningful latent space.

\section{CONCLUSION}\label{sec:conclusion}
EBMs have recently demonstrated strong performance across diverse generative modelling tasks, including image and language modelling \citep{thornton2025controlled,wang2025equilibrium, gladstone2025}. They offer distinctive capabilities that are difficult to achieve with competing approaches -- most notably, direct anomaly detection by thresholding the energy, and compositional generation \citep{thornton2025controlled, du2020compositional} by combining multiple independently trained \glspl*{ebm} at inference time. Despite these appealing properties, they remain hard to train due to their double intractable nature.

Motivated by this and the recent advances in particle-based training methods, we propose an interacting particle method for learning latent energy-based models (\gls*{iplebm}), which provides a more computationally efficient method to train \glspl*{lebm}. By discretising a system of SDEs that is designed to admit an invariant measure concentrating around the maximiser of the marginal log-likelihood, we obtain a practical algorithm that is both scalable and theoretically grounded (Section \ref{sec:algorithm}). In Section \ref{sec:nonasymptotic}, we provide to the best of our knowledge the first convergence bounds for training latent energy-based models under the strong log-concavity and smoothness assumptions. In particular,   we show that the algorithm converges to the maximiser of the empirical marginal log-likelihood with increasing number of both data points and particles; this ensures that even with a small number of particles reliable performance can be achieved on large datasets. (Theorem \ref{thm:exact-gradient}, Theorem \ref{thm:inexact-gradient}). Although the convergence bounds are obtained under the strong log-concavity we expect those can be further relaxed~\citep{zhang2023nonasymptotic} and highlight that the particle system is designed to concentrate around the empirical maximisers of the marginal likelihood even in the non log-concave settings.

Lastly, while for fairness of comparison we use a simple Langevin chain to approximate the prior expectation, the framework is agnostic to the choice of sampler: advanced samplers such as Hamiltonian Monte Carlo (HMC) \citep{duane1987hybrid}
could be employed to improve the generative performance. In addition, extensions to underdamped settings offer a promising direction to obtain accelerated versions of \gls*{iplebm} with theoretical guarantees  \citep{oliva2024}.

\acknowledgements{
J.~M. is supported by EPSRC through the Modern Statistics and Statistical Machine Learning (StatML) CDT programme, grant no. EP/Y034813/1. T.~W. is supported by the Roth Scholarship from the Department of Mathematics, Imperial College London. We acknowledge computational resources and support provided by the Department of Mathematics and the Imperial College Research Computing Service, DOI: 10.14469/hpc/2232.
}

\bibliographystyle{apalike}
\bibliography{references}

\onecolumn
\title{Learning Latent Energy-Based Models via Interacting Particle Langevin Dynamics \\
Supplementary Material}
\maketitle

\appendix
\section{Preliminary results}\label{app:gradient-derivations}
Recall that for a data point $y$, we define $p_\theta(x,y) = p_\alpha(x)p_\beta(y \mid x)$, and we use the notation $\phi_y(\theta, x)$:
\begin{equation}
    \phi_{y}(\theta, x) = - \log p_\theta(x, y).
\end{equation}
We can write the gradient:
\begin{equation}
     \nabla_\theta  \log p_{\theta}(x,y) = \nabla_\theta \log p_\alpha(x) + \nabla_\theta \log p_\beta (y \mid x).
\end{equation}

\subsection{Derivation of $\nabla_\alpha \phi_y(\theta, x)$}
 Since $p_\beta(y_m |x)$ is independent of $\alpha$ the gradient reads 
\begin{equation}
    \nabla_\alpha \phi_y(\theta, x) = -\nabla_\alpha\log p_\theta(x, y) = -\nabla_\alpha \log p_\alpha(x).
\end{equation}

By definition $p_\alpha(x) = e^{-U_\alpha(x)}/Z_\alpha$, so
\begin{equation}
    \nabla_\alpha \phi_y(\theta, x) = -\nabla_\alpha \log p_\alpha(x) = \nabla_\alpha U_\alpha(x) + \nabla_\alpha \log Z_\alpha 
\end{equation}
where $Z_\alpha =  \int e^{-U_\alpha(x)} \dd x$ is the normalising constant which is unknown. However, provided differentiation under the integral sign is justified the gradient of  $\log Z_\alpha$ can be rewritten as
\begin{align}
    &\nabla_\alpha \log Z_\alpha = \frac{\nabla_\alpha Z_\alpha}{Z_\alpha} = \frac{\int \nabla_\alpha e^{-U _\alpha (x)} \dd x}{Z_\alpha} \\
    &=  - \int \nabla_\alpha U_\alpha(x) p_\alpha(x) \dd x = - \mathbb{E}_{p_\alpha(x)}[\nabla_\alpha U_\alpha(x)]
\end{align}
hence
\begin{align}
   \nabla_\alpha \phi_y(\theta, x) =  \nabla_\alpha U_\alpha(x) - \mathbb{E}_{p_\alpha(x)}[\nabla_\alpha U_\alpha(x)]
\end{align}

and for a set of particles $\{X_t^{m,n}\}_{m=1, n=1}^{m=M, n=N}$ and a parameter $\theta_k = (\alpha_k, \beta_k)$

\begin{align}
    \frac{1}{MN} \sum_{m=1}^M \sum_{n=1}^N  \nabla_{\alpha} \phi_{y_m}(\theta_k, X_k^{m,n}) = \frac{1}{MN} \sum_{m=1}^M \sum_{n=1}^N \nabla_\alpha U_{\alpha_k}(X_k^{m,n}) - \mathbb{E}_{p_{\alpha_k}(x)}[\nabla_\alpha U_{\alpha_k}(x)]
\end{align}

\subsection{Derivation of $\nabla_\beta \phi_y(\theta, x)$}

Similarly, given a data point $y$ since $p_\alpha(x)$ is independent of $\beta$ the gradient of $\phi_y (\theta, x)$ with respect to $\beta$ reads

\begin{equation}
    \nabla_\beta \phi_y(\theta, x) = - \nabla_\beta\log p_\theta(x, y) = -\nabla_\beta \log p_\beta(y | x)
\end{equation}

Let $V_\beta(x, y) = -\log p_\beta(y \mid x)$. Then for a set of particles $\{X_t^{m,n}\}_{m=1, n=1}^{m=M, n=N}$ and a parameter $\theta_k = (\alpha_k, \beta_k)$

\begin{equation}
    \frac{1}{MN} \sum_{m=1}^M \sum_{n=1}^N  \nabla_{\beta} \phi_{y_m}(\theta_k, X_k^{m,n}) = \frac{1}{MN} \sum_{m=1}^M \sum_{n=1}^N  \nabla_{\beta} V_\beta(X_k^{m,n},y_m))
\end{equation}

which, if we assume a standard isotropic Gaussian decoder, that is, $p(y_m \mid x)=\mathcal{N}(y_m; g_\beta(x), \sigma^2 I)$, then $V_\beta(x,y_m) = (1/{2\sigma^2}) {\|y_m - g_{\beta}(x)\|^2} + \text{const}$, where $g_\beta$ denotes the decoder (generator).
\subsection{Derivation of $\nabla_x \phi_y(\theta, x)$}

Recall $V_{\beta}(x, y) = - \log p_\beta(y \mid x)$. Then given a data point $y$, the gradient of $\phi_y (\theta, x)$ with respect to $x$ reads

\begin{align}\label{eq:x_gradient}
     \nabla_{x} \phi_y (\theta, x) &= -\nabla_{x}  \log p_{\theta}(x,y) = -\nabla_{x} \log p_\alpha(x) - \nabla_x \log p_\beta (y|x) \\
     &=\nabla_{x} U_\alpha(x) + \nabla_{x} V_\beta(x, y)
\end{align}

since the normalising constant $Z_\alpha$ is independent of $x$.

\section{Proofs}
\subsection{Proof of Theorem \ref{thm:exact-gradient}}\label{sec_app:proof-exact-gradient}
\begin{proof}
Recall the discretised system of SDEs defined in \eqref{eq:ipla-algo-1}--\eqref{eq:ipla-algo-2}
\begin{align}
\theta_{k+1} &= \theta_k - \frac{h}{MN} \sum_{m=1}^M \sum_{n=1}^N \nabla_{1} \phi_{y_m}(\theta_k, X_k^{m,n}) + \sqrt{\frac{2h}{MN}} W_k, \label{app:eq:ipla-algo-1-proof} \\
X_{k+1}^{m,n} &= X_k^{m,n} - h \nabla_{2} \phi_{y_m}(\theta_k, X_k^{m,n}) + \sqrt{2h} W_k^{m,n}, \label{app:eq:ipla-algo-2-proof}
\end{align}
where $\phi_{y_m}(\theta, x) = -\log p_\theta(x, y_m)$ and which is equivalent to the SDE system in \eqref{eq:alpha-update}--\eqref{eq:x-update}. In the above display, we use $\nabla_1$ and $\nabla_2$ to denote the derivatives w.r.t. first and second arguments of $\phi_{y_m}$, respectively, as this will be important in the proof.

Consider the function $\Phi: \mathbb{R}^{d_\theta} \times (\mathbb{R}^{d_x})^{M \times N} \to \mathbb{R}$, defined as
\begin{align*}
\Phi(\theta, z^{1:M, 1:N}) = \sum_{m=1}^M \sum_{n=1}^N \phi_{y_m}(\theta, \sqrt{MN} z^{m,n}).
\end{align*}
Next, let us define the following probability distribution:
\begin{align*}
\pi(\theta, z^{1:M, 1:N}) \propto \exp(-\Phi(\theta, z^{1:M, 1:N})).
\end{align*}
Our strategy in this proof is to notice that the iterates \eqref{app:eq:ipla-algo-1-proof}--\eqref{app:eq:ipla-algo-2-proof} can be seen as a discretisation of an SDE, targeting $\pi$. In order to see this, consider the Langevin SDE targeting $\pi$:
\begin{align}
\mathrm{d} \theta_t &= - \sum_{m=1}^M \sum_{n=1}^N \nabla_1 \phi_{y_m}(\theta_t, \sqrt{MN} Z_t^{m,n}) \mathrm{d}t + \sqrt{2} \mathrm{d}B_t, \label{app:eq:ipla-scaled-sde-1}\\
\mathrm{d} Z_t^{m,n} &= - \sqrt{MN} \nabla_2 \phi_{y_m}(\theta_t, \sqrt{MN} Z_t^{m,n}) \mathrm{d}t + \sqrt{2} \mathrm{d}B_t^{m,n}. \label{app:eq:ipla-scaled-sde-2}
\end{align}
Euler-Maruyama discretisation of this SDE with step-size $\bar{h} = h/MN$ gives us
\begin{align}
\theta_{k+1} &= \theta_k - \frac{h}{MN} \sum_{m=1}^M \sum_{n=1}^N \nabla_1 \phi_{y_m}(\theta_k, \sqrt{MN} Z_k^{m,n}) + \sqrt{\frac{2h}{MN}} W_k, \label{app:eq:ipla-scaled-1} \\
Z_{k+1}^{m,n}&= Z_k^{m,n} - \frac{h}{\sqrt{MN}} \nabla_2 \phi_{y_m}(\theta_k, \sqrt{MN} Z_k^{m,n}) + \sqrt{\frac{2h}{MN}} W_k^{m,n}. \label{app:eq:ipla-scaled-2}
\end{align}
Now, if we define $X_k^{m,n} = \sqrt{MN} Z_k^{m,n}$, we recover exactly \eqref{eq:ipla-algo-1}--\eqref{eq:ipla-algo-2} (i.e. \eqref{app:eq:ipla-algo-1-proof}--\eqref{app:eq:ipla-algo-2-proof} above). In other words, up to a simple rescaling, the systems are identical. Most importantly, $\theta$-marginals of both scaled algorithms are identical, thus the analysis of the iterates defined in \eqref{app:eq:ipla-scaled-1} will give us the convergence of the parameters.

To assess first whether this target distribution is desirable, we consider the $\theta$-marginal of $\pi$, denoted by $\pi_\Theta$. That is, $\pi_\Theta$ is obtained by integrating out all latent variables $z^{1:M,1:N}$ from the joint distribution $\pi$: 
\begin{align*}
\pi_\Theta(\theta) &= \int \pi(\theta, z^{1:M, 1:N}) \mathrm{d}z^{1:M, 1:N} =\int \exp(-\Phi(\theta, z^{1:M, 1:N})) \mathrm{d}z^{1:M, 1:N} \\
&= \int \prod_{m=1}^M \prod_{n=1}^N \exp(-\phi_{y_m}(\theta, \sqrt{MN} z^{m,n})) \mathrm{d}z^{1:M, 1:N} = \prod_{m=1}^M \prod_{n=1}^N \int \exp(-\phi_{y_m}(\theta, \sqrt{MN} z^{m,n})) \mathrm{d}z^{m,n}, \\
&\propto \prod_{m=1}^M \left(\int \exp(-\phi_{y_m}(\theta, x)) \mathrm{d}x\right)^N \propto \prod_{m=1}^M p_\theta(y_m)^N,\\
&\propto \exp\left(N \sum_{m=1}^M \log p_\theta(y_m)\right).
\end{align*}
Thus, for a fixed dataset $\{y_m\}_{m=1}^M$ as $N \to \infty$, the distribution $\pi_\Theta(\theta)$ concentrates around the maximisers of the target loss $\ell_M$ and we will aim to quantify this concentration rate. To proceed, we first establish the convexity properties of the function $\Phi$. Under \Cref{ass:strong-convexity} and \Cref{ass:L-smooth}, $\phi_{y_m}$ is $\mu$-strongly convex and $L$-smooth in both arguments for all $m \in \{1,\ldots,M\}$.
By arguments analogous to those in \citet[Lemmas~A.4--A.5]{oliva2024}, it follows that $\Phi$ is $MN\mu$-strongly convex and $MNL$-smooth in both arguments. Since $-\log \pi_\Theta(\theta) = -N \sum_{m=1}^M \log p_\theta(y_m) + \mathrm{const}$, and each $-\log p_\theta(y_m)$ is $\mu$-strongly convex in $\theta$ by Remark~\ref{remark:strong-convexity} $\pi_\Theta$ is $\mu M N$-strongly log-concave. Hence, to obtain the concentration rate we can apply \citet[Lemma~A.8]{altschuler2024faster}, which gives 
\begin{align*}
W_2(\delta_{\theta_\star}, \pi_\Theta) \leq \sqrt{\frac{d_\theta}{\mu MN}}.
\end{align*}

Next, we look at the rate of convergence. Since $\Phi$ is $MN\mu$-strongly convex and $MNL$-smooth in both arguments, an exponential convergence rate will hold by using well-known convergence results of Langevin diffusions. Let $\nu_k = \mathcal{L}(\theta_k, Z_k^{1:M, 1:N})$ be the law of the system in \eqref{app:eq:ipla-scaled-1}--\eqref{app:eq:ipla-scaled-2}. Given that, the target $\pi$ is $\bar{\mu} = MN\mu$-strongly log-concave and $\bar{L} = M N L$ log-smooth, the discretisation of this system with $\bar{h} = h /MN$ satisfies \citep[Theorem~1]{dalalyan2019user}
\begin{align}
W_2(\nu_k, \pi) \leq (1 - \bar{h} \bar{\mu})^k W_2(\nu_0, \pi) + 1.65 \left(\frac{\bar{L}}{\bar{\mu}}\right) \sqrt{{d_\theta + M N d_x}} \bar{h}^{1/2},
\end{align}
where $0 < \bar{h} \leq 2 / (\bar{\mu} + \bar{L})$. Plugging $\bar{h} = h / MN$, $\bar{\mu} = MN \mu$, and $\bar{L} = MNL$ to the above bound, we get
\begin{align}
W_2(\nu_k, \pi) \leq (1 - {h} {\mu})^k W_2(\nu_0, \pi) + 1.65 \left(\frac{{L}}{{\mu}}\right) \sqrt{\frac{d_\theta + M N d_x}{MN}} h^{1/2},
\end{align}
where the step-size restriction can also be simplified to $0 < h \leq 2 / (\mu + L)$. Now, let us denote $\nu^\theta_k = \mathcal{L}(\theta_k)$ and $\pi_\Theta(\theta) = \int \pi(\theta, z^{1:M, 1:N}) \mathrm{d}z^{1:M, 1:N}$, then by the definition of Wasserstein distance, we have
\begin{align*}
W_2(\nu_k^\theta, \pi_\Theta) &\leq W_2(\nu_k, \pi),
\end{align*}
concluding the convergence part.

Merging these results, we arrive at
\begin{align}
\mathbb{E}\left[\|\theta_k - \theta_\star\|^2\right]^{1/2} \leq (1-\mu h)^k W_2(\nu_0, \pi) + 1.65 \left(\frac{L}{\mu}\right) \sqrt{\frac{d_\theta + M N d_x}{MN}} h^{1/2} + \sqrt{\frac{d_\theta}{\mu MN}}.
\end{align}
\end{proof}

\subsection{Preliminary result for Theorem~\ref{thm:inexact-gradient}}
In this section, we rework Theorem~4 in \citet{dalalyan2019user} in the same assumption setting as in \Cref{ass:bound-on-bias}. Let $f: \mathbb{R}^p \to \mathbb{R}$ be a $\mu$-strongly convex function with $L$-Lipschitz gradients. Let $\boldsymbol{\vartheta}_{k, h}$ denote the iterates of the noisy LMC (nLMC) algorithm as defined in \citet{dalalyan2019user}
\begin{equation}
    \boldsymbol{\vartheta}_{k+1, h}=\boldsymbol{\vartheta}_{k, h}-h \nabla f\left(\boldsymbol{\vartheta}_{k, h}\right)+h \boldsymbol{\zeta}_k+\sqrt{2 h} \boldsymbol{\xi}_{k+1} ; \quad k=0,1,2, \ldots
\end{equation}

where $h>0$ and $\boldsymbol{\xi}_{k+1}$ are a collection of standard Gaussian random variables of appropriate dimension.
Suppose for some $\delta>0$ and $\sigma>0$ and for every $k \in \mathbb{N}$, \Cref{ass:bound-on-bias} holds for $(\boldsymbol{\zeta}_k)_{k\geq 0}$. We then have the following proposition.

\begin{proposition}[Theorem 4 in \citet{dalalyan2019user} adapted]\label{app:prop:dalalyan}
 Let $\boldsymbol{\vartheta}_{K, h}$ be the $K$'th the iterate of the nLMC algorithm  and $v_K$ be its distribution. If the function $f:\mathbb{R}^p \to \mathbb{R}$ is $L$-smooth and $\mu$-strongly convex and $h \leq 2 /(\mu + L)$ then
$$
\begin{aligned}
& W_2\left(v_K, \pi\right) \leq(1- \mu h)^K W_2\left(v_0, \pi\right)+ 1.65(L / \mu)(h p)^{1 / 2}  +\frac{\delta }{\mu}+\frac{\sigma^2h^{1 / 2}}{1.64 L p^{1/2}+\sigma \sqrt{\mu}}
\end{aligned}
$$
\end{proposition}

\begin{proof}
    Under \Cref{ass:bound-on-bias}, modified Proposition 2 in \citet{dalalyan2019user} yields
\begin{equation}
    W_2\left(\nu_{k+1}, \pi\right)^2 \leq\left\{(1- \mu h) W_2\left(\nu_k, \pi\right)+\alpha L\left(h^3 p\right)^{1 / 2}+h \delta\right\}^2+\sigma^2 h^2
\end{equation}
Applying \citet[Lemma~1]{dalalyan2019user} with $A=\mu h, B=\sigma h $ and $C=\alpha L \left(h^3 p\right)^{1 / 2}+h \delta$, implies that $W_2\left(\nu_k, \pi\right)$ is less than or equal to
\begin{equation}
    (1-\mu h)^k W_2\left(\nu_0, \pi\right)+\frac{\alpha L(h p)^{1 / 2}+\delta} {\mu}+\frac{\sigma^2 h }{\alpha L (hp)^{1 / 2}+\delta+(\mu h)^{1 / 2} \sigma}
\end{equation}

where $ \alpha  = 7\sqrt{2}/6 \in (1.64, 1.65)$.

\end{proof}

\subsection{Proof of Theorem \ref{thm:inexact-gradient}}\label{sec_app:proof-inexact-gradient}

\begin{proof}
Let us define $\tilde{\theta}_k = (\tilde{\alpha}_k, \tilde{\beta}_k)$. We can compactly write \eqref{eq:alpha-update-inexact}--\eqref{eq:x-update-inexact} as:
\begin{align}
\tilde{\theta}_{k+1} &= \tilde{\theta}_k - \frac{h}{MN} \sum_{m=1}^M \sum_{n=1}^N \nabla_{1} \phi_{y_m}(\tilde{\theta}_k, \tilde{X}_k^{m,n}) + h\boldsymbol{\zeta}_k + \sqrt{\tfrac{2h}{MN}} W_k, \label{app:eq:ipla-algo-1-noisy-proof} \\
\tilde{X}_{k+1}^{m,n} &= \tilde{X}_k^{m,n} - h \nabla_{2} \phi_{y_m}(\tilde{\theta}_k, \tilde{X}_k^{m,n}) + \sqrt{2h} W_k^{m,n},\label{app:eq:ipla-algo-2-noisy-proof}
\end{align}
where $\mathbb{R}^{d_\theta} \ni \boldsymbol{\zeta}_k = (\zeta_k, 0_{d_\beta})^\intercal$ and where $\zeta_k:=g_k-\mathbb{E}_{p_{\tilde\alpha_k}(x)}[\nabla_\alpha U_{\tilde\alpha_k}(x)]$ and $g_k:=M^{-1}\sum_{m=1}^M\nabla_\alpha U_{\tilde\alpha_k}(\hat X^m_{k,J})$.
Following the same steps as in the proof of Theorem~\ref{thm:exact-gradient} we rescale the system with step-size $\tilde{h} = \frac{h}{MN}$. Note that this results in a $\tilde{h}MN\zeta_k$ term in Equation~\eqref{app:eq:ipla-algo-1-noisy-proof} and that by \Cref{ass:bound-on-bias}  we have

\begin{align*}
        \mathbb{E}\left[\left\|\mathbb{E}\left[MN\zeta_k | \mathcal{F}_k \right]\right\|^2\right] &\le (MN\delta)^2, \\
        \mathbb{E}\left[\left\|MN\zeta_k - \mathbb{E}\left[MN\zeta_k | \mathcal{F}_k \right]\right\|^2\right] &\le (MN\sigma)^2
    \end{align*}

Using $\tilde{h} = \frac{h}{MN}$, $\tilde{\mu} = MN \mu$, $\tilde{L} = MNL$, $\tilde{\delta} = MN \delta$ and $\tilde{\sigma} = MN \sigma$ to Proposition~\ref{app:prop:dalalyan} we obtain the following bound:

\begin{align}\label{app:eq:inexact-gradient-bound-new}
\mathbb{E}\left[\|\tilde{\theta}_k - \theta_\star\|^2\right]^{1/2} &\leq (1-\mu h)^k W_2(\nu_0, \pi) + 1.65 \left(\frac{L}{\mu}\right)\sqrt{\frac{d_\theta + M N d_x}{MN}} h^{1/2} \nonumber \\
&+ \frac{\sigma^2\sqrt{h}}{1.64 L  \sqrt{\frac{d_\theta + MN d_x}{MN}}+ \sigma \sqrt{\mu}} + \frac{\delta}{\mu} + \sqrt{\frac{d_\theta}{\mu MN}}.
\end{align}

We can then identify
\begin{align*}
\tilde{C}_0 &= W_2(\nu_0, \pi), \\
\tilde{C}_1 &= 1.65 \left(\frac{L}{\mu}\right)\sqrt{\frac{d_\theta + M N d_x}{MN}} + \frac{\sigma^2}{1.64 L  \sqrt{\frac{d_\theta + MN d_x}{MN}}+ \sigma \sqrt{\mu}}, \\
\tilde{C}_2 &= \sqrt{\frac{d_\theta}{\mu}}, \\
\tilde{C}_3 &= \frac{\delta}{\mu}
\end{align*}
\end{proof}

\section{Implementation Details}
\subsection{A Practical Algorithm}\label{app:practical-algorithm}
We now provide a few modifications to enable efficient simulation of the gradient flow. We include the pseudocode in \Cref{algo:practical}.

\paragraph{Subsampling} The updates in Eqs.~\eqref{eq:alpha-update-inexact}-\eqref{eq:x-update-inexact}
are prohibitively expensive due to the need of averaging over the entire training set of $M$ data points. Instead, we adopt the mini-batching strategy \citep{kuntz2023,wang2025} and compute a subsampled version of the losses. For a minibatch of indices $\mathcal{B} \subseteq [M]$, we compute the energy loss and generator loss as:
\begin{align}
\hat{\mathcal{L}}_{d}(X^{1:M,1:N}, \mathcal{B}) & = \frac{1}{N|\mathcal{B}|} \sum_{(m,n)\in \mathcal{B}\times [N]} V(X^{m,n}, y^m),\label{eq:subsampled_generator} \\
\hat{\mathcal{L}}_e (\hat{X}_J^{1:M}, X^{1:M, 1:N},\mathcal{B}) & = \frac{1}{N|\mathcal{B}|} \sum_{(m,n)\in \mathcal{B} \times [N]} U(X^{m,n}) - \frac{1}{|\mathcal{B}|} \sum_{m\in\mathcal{B}} U(\hat{X}_J^{m}),\label{eq:subsampled_energy}
\end{align}
where $X^{1:M,1:N}$ are the posterior particles, $\hat{X}_J^{1:M}$ are particles sampled from the energy prior using $J$ Langevin steps. This reduces the computational cost for each update to the posterior particles from $\mathcal{O}(MN)$ to $\mathcal{O}(|\mathcal{B}|N)$.
\paragraph{Noise Addition} Due to the use of mini-batching, we also adapt the Langevin dynamics to account for the noise added to the posterior particles $X^{1:M,1:N}$. For $M=BL$, where $B=|\mathcal B|$, so that one epoch contains $L$ mini-batch updates and each cloud $X^{m,1:N}$ receives one posterior drift update per epoch. We add the Brownian motion scaled by $\sqrt{1/L}$ to all particles every time the generator is updated, which matches with the amount of noise added in~\eqref{eq:x-update-inexact}. When $M$ is not divisible by $B$, we set $L=M/B$ as an effective noise-scaling factor and sample mini-batches according to the implementation described below. Since the noise addition does not involve gradient computations, it is relatively cheap compared to the posterior updates.

\paragraph{Adaptive Optimisers} To accelerate convergence, we use the Adam optimiser \cite{kingma2017adammethodstochasticoptimization} for the network parameters $(\alpha, \beta)$, following the practice in \citet{pang2020ebmprior,schroeder2023energy}. We detail the hyperparameters in \Cref{app:synthetic-experiment-details} and \Cref{app:image_details}.

\begin{algorithm}[htbp]
    \caption{A practical version of \gls*{iplebm}}
    \label{algo:practical}
    \begin{algorithmic}[1]
    \Require Number of training iterations $K$, number of particles $N$, number of prior iterations $J$, initial parameters $(\alpha_0, \beta_0)$, observed data $\{y^m\}_{m=1}^M$, batch size $|\mathcal{B}|$, scaling $L=M/|\mathcal{B}|$, initial particles $X_0^{1:M, 1:N}$, stepsizes $h > 0 $ and $\gamma > 0$
    \For{$k = 0$ to $K-1$}
        \State Sample batch of indices $\mathcal{B} \subseteq [M]$
        \State $X_{k+1}^{1:M,1:N} \gets X_k^{1:M,1:N}$
        \State $X_{k+1}^{\mathcal{B},1:N} \gets X_{k}^{\mathcal{B},1:N} - h \left[\nabla_x U_{\alpha_k}(X_{k}^{\mathcal{B},1:N}) + \nabla_x V_{\beta_k}(X_{k}^{\mathcal{B},1:N}, y^{\mathcal{B}})\right]$
        \Comment{Update posterior}
        \State $X_{k+1}^{1:M,1:N} \gets X_{k+1}^{1:M,1:N} + \sqrt{{2h}/{L}} \, \tilde{W}_k^{1:M,1:N}$
            \State \textcolor{black}{Sample $\hat{X}_{k,0}^{\mathcal{B}} \sim \mathcal{N}(0,I)$}
            \For{$j=0$ to $J-1$}
            \Comment{Generate new auxiliary prior samples}
                \State $\hat{X}_{k,j+1}^{\mathcal{B}} \gets \hat{X}_{k,j}^{\mathcal{B}} - \gamma \, \nabla_x U_{\alpha_k}(\hat{X}_{k,j}^{\mathcal{B}}) + \sqrt{2\gamma} \, W_{k,j}^{\mathcal{B}}$
            \EndFor
        \State \textcolor{black}{Compute energy loss $\hat{\mathcal{L}}_{e}=\hat{\mathcal{L}}_e (\hat{X}_{k,J}^{\mathcal{B}}, X_{k}^{1:M,1:N},\mathcal{B})$ in (\ref{eq:subsampled_energy})}
        \State \textcolor{black}{Compute generator loss $\hat{\mathcal{L}}_d=\hat{\mathcal{L}}_{d}(X_{k}^{1:M,1:N},\mathcal{B})$ in (\ref{eq:subsampled_generator})}
        \State $\alpha_{k+1} \gets \mathrm{OptimizerStep(\alpha_k, \hat{\mathcal{L}}_e)}$
        \State $\beta_{k+1}\gets \mathrm{OptimizerStep}(\beta_k, \hat{\mathcal{L}}_d)$
    \EndFor
    \end{algorithmic}
\end{algorithm}

\subsection{EBIPLA with Warmup for Image Experiments}\label{app:warm_images}
To quickly traverse through the transient phase for the posterior particles in image experiments \citep{kuntz2023}, we adopt a warm-up scheme similar to the predictor-corrector SDE solver in \citet{song2021score}, correcting the solution to the posterior SDE in \eqref{eq:IPLD_2} using Langevin dynamics. We apply the warm-up during the first $S_{\mathrm{warm}}$ training iterations and run the Langevin dynamics for $10$ steps at each warm-up iteration. After the warm-up, we initialise $X_{\mathrm{post}}^{m,1:N}$ by creating $N$ identical copies of each $X_{\mathrm{post}}^{m,1}$ obtained from the persistent chains. The Gaussian noises in the inner warm-up Langevin steps are sampled independently across inner iterations. We provide the pseudocode in \Cref{algo:practical_warm}.
\begin{algorithm}[h]
    \caption{A practical version of \gls*{iplebm} with warmup}
    \label{algo:practical_warm}
    \begin{algorithmic}[1]
    \Require Number of training iterations $K$,  number of prior iterations $J$, warmup iterations $S_{\text{warm}}$, initial parameters $(\alpha_0, \beta_0)$, observed data $\{y^m\}_{m=1}^M$, batch size $|\mathcal{B}|$, scaling $L=M/|\mathcal{B}|$, initial particles $X_0^{1:M, 1:N'}$ with $N'$, final number of particles $N$ with $N\equiv 0 \ (\text{mod} \ N')$,  stepsizes $h > 0$, $h_{\mathrm{warm}} > 0$, and $\gamma > 0$.
    \For{$k = 0$ to $K-1$}
        \State Sample batch of indices $\mathcal{B} \subseteq [M]$
        \Comment{Update posterior samples}
        \State \textcolor{black}{$X_{k+1}^{1:M,1:N'} \gets X_k^{1:M,1:N'}$}
        \If{\textcolor{black}{$k< S_{\text{warm}}$}}
            \State \textcolor{black}{$X^\dagger \gets X_{k}^{\mathcal{B},1:N'}$}
            \For{$i=1, \dots, 10$}
                \State \textcolor{black}{$X \gets X^\dagger - h_{\mathrm{warm}} \left[\nabla_x U_{\alpha_k}(X^\dagger) + \nabla_x V_{\beta_k}(X^\dagger, y^\mathcal{B})\right] + \sqrt{2h_{\text{warm}}}\epsilon_{k,i}^{\mathcal{B}}$}
            \EndFor
            \State \textcolor{black}{$X_{k+1}^{\mathcal{B},1:N'} \gets X^\dagger$}
        \ElsIf{\textcolor{black}{$k=S_{\text{warm}}$}}
            \State \textcolor{black}{Replicate each particle cloud in $X_{k+1}^{1:M,1:N'}$ to obtain $X_{k+1}^{1:M,1:N}$}
            \State \textcolor{black}{Set $N'\gets N$}
        \Else
            \State \textcolor{black}{$X_{k+1}^{\mathcal{B},1:N'} \gets X_{k}^{\mathcal{B},1:N'} - h \left[\nabla_x U_{\alpha_k}(X_{k}^{\mathcal{B},1:N'}) + \nabla_x V_{\beta_k}(X_{k}^{\mathcal{B},1:N'}, y^\mathcal{B})\right]$}
            \State \textcolor{black}{$X_{k+1}^{1:M,1:N'} \gets X_{k+1}^{1:M,1:N'} + \sqrt{{2h}/{L}} \, \tilde{W}_k^{1:M,1:N'}$}
        \EndIf
        \State \textcolor{black}{Sample $\hat{X}_{k,0}^{\mathcal{B}} \sim \mathcal{N}(0,I)$}
            \For{$j=0$ to $J-1$}
            \Comment{Generate new auxiliary prior samples}
                \State \textcolor{black}{$\hat{X}_{k,j+1}^{\mathcal{B}} \gets \hat{X}_{k,j}^{\mathcal{B}} - \gamma \, \nabla_x U_{\alpha_k}(\hat{X}_{k,j}^{\mathcal{B}}) + \sqrt{2\gamma} \, W_{k,j}^{\mathcal{B}}$}
            \EndFor
        \State \textcolor{black}{Compute energy loss $\hat{\mathcal{L}}_{e}=\hat{\mathcal{L}}_e (\hat{X}_{k,J}^{\mathcal{B}}, X_{k+1}^{1:M,1:N'},\mathcal{B})$ in (\ref{eq:subsampled_energy})}
        \State \textcolor{black}{Compute generator loss $\hat{\mathcal{L}}_d=\hat{\mathcal{L}}_{d}(X_{k+1}^{1:M,1:N'},\mathcal{B})$ in (\ref{eq:subsampled_generator})}
        \State $\alpha_{k+1} \gets \mathrm{OptimizerStep(\alpha_k, \hat{\mathcal{L}}_e)}$
        \State $\beta_{k+1}\gets \mathrm{OptimizerStep}(\beta_k, \hat{\mathcal{L}}_d)$
    \EndFor
    \end{algorithmic}
\end{algorithm}
\subsection{Computational Resources}
All experiments were conducted on internal HPCs with NVIDIA L40S (48G) GPUs, internal servers with NVIDIA RTX 3090 (24G) GPUs, or rented cloud GPUs. 

\section{Experimental Details}\label{app:experiments}

\subsection{Synthetic Experiments}\label{app:synthetic-experiment-details}

For implementation of \gls*{iplebm} on the synthetic dataset we use the practical algorithm as detailed in Section \ref{app:practical-algorithm}. For fairness of comparison we implement \gls*{lebm} \citep{pang2020ebmprior} without the prior tilting introduced by the authors. We observed that for our example dataset the tilting which is a form of prior regularisation hindered the performance of both \gls*{iplebm} and \gls*{lebm} and hence decided to drop it for both methods.

\paragraph{Architecture} 
 We set the standard deviation of the Gaussian decoder for both models to $\sigma=0.05$. For the prior energy \(U_\alpha(x)\) we use a multi-layer perceptron (MLP) with three hidden layers, each with 128 hidden units and SiLU activation for \gls*{iplebm} and ReLU activation for \gls*{lebm}. We observed that ReLU worked better for the baseline method. The generator \(g_\beta(z):\mathbb{R}^{d_z}\!\to\!\mathbb{R}^{d_x}\) is a single linear layer mapping.

\paragraph{Training} We train both models for 200 epochs with the batch size of 1000 data points. For the parameter updates $(\alpha, \beta)$ we use the Adam optimiser with learning rate $1 \times10^{-2}$ for both. The exponential decay rates for the first and second moment estimates were configured as $\beta_1 = 0.5$ and $\beta_2 = 0.999$, respectively. We vary the prior \gls*{ula} step size $\gamma$ and the posterior discretisation step size $h$ in \gls*{iplebm}, as well as the prior $\gamma$ and posterior $h$ \gls*{ula} step sizes in \gls*{lebm}, according to the number of posterior particles or MCMC steps (see Table~\ref{tab:ebipla-stepsizes}, Table~\ref{tab:lebm-stepsizes}).

\subsubsection{Dataset creation details}\label{app:additional-synthetic-results}

\begin{figure}
    \centering
    \includegraphics[width=0.45\linewidth]{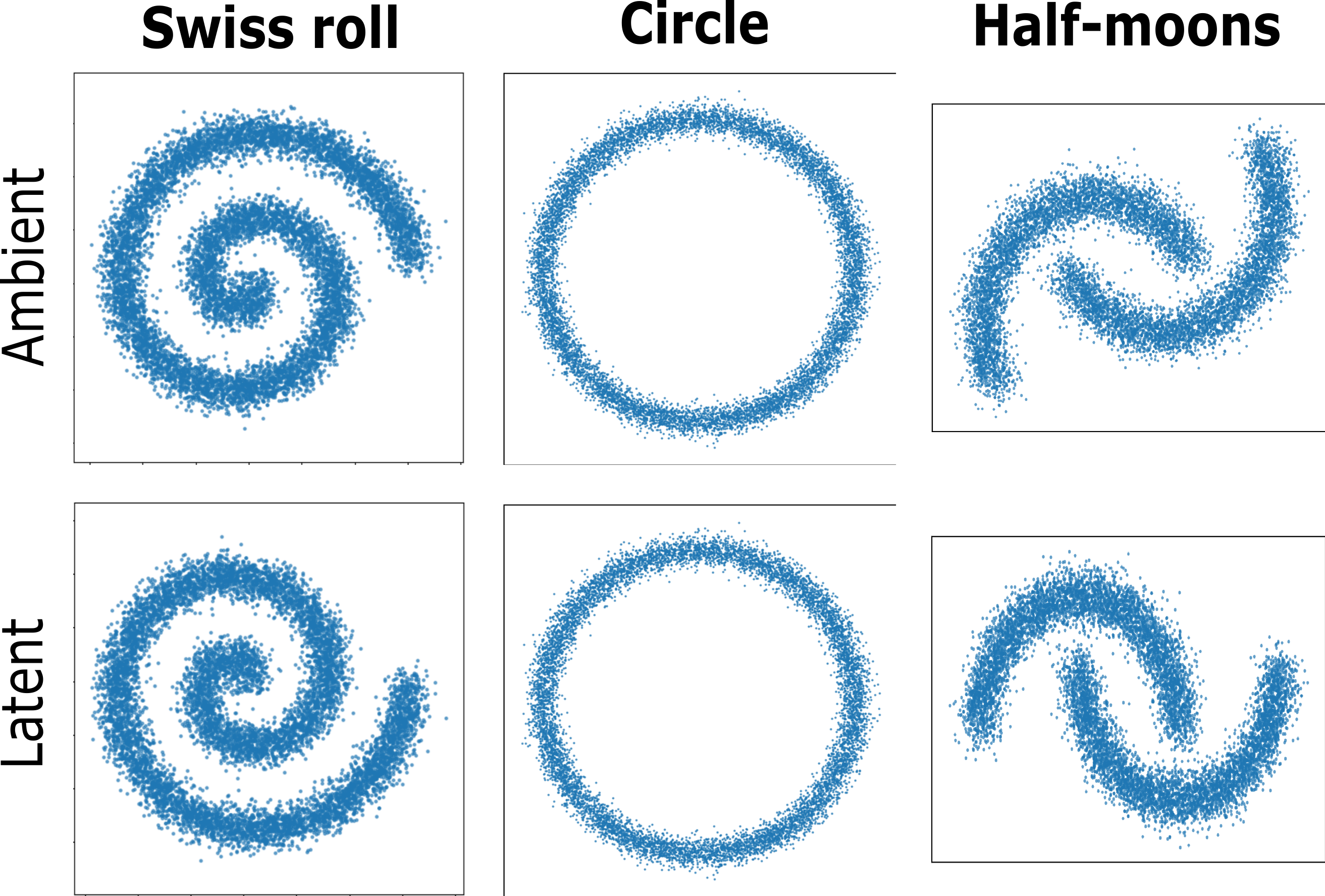}
    \caption{Visualisation of the datasets in latent and ambient space}
    \label{fig:data-visualisation}
\end{figure}

\paragraph{Swiss roll} 
To create the rotated Swiss roll we sample \(t \in [t_{\min}, t_{\max}]\) uniformly in arc length to ensure even distribution of the data points. Latent points are $x_{\text{lat}}(t)=\tfrac{1}{8}\big[t\cos t,\; t\sin t\big]+\varepsilon,\quad \varepsilon\sim\mathcal{N}(0,0.01I).$ We map to the ambient space via a fixed linear decoder \(T: \R^2 \to \R^2\) with orthonormal rows, which results in a rotated Swiss roll (see Figure \ref{fig:data-visualisation}). The dataset size was set to 10,000 data points.
\paragraph{Half-moons}

To create the rotated Half-moons dataset, we use the \texttt{make\_moons} function from \texttt{sklearn} to sample $N=10000$ points from the standard two-moons distribution in latent space $\mathbb{R}^2$ with zero intrinsic noise. We then add isotropic Gaussian perturbations $x_{\text{lat}} = x_{\text{moon}} + \varepsilon,  \varepsilon \sim \mathcal{N}(0,\,0.01I)$
which corresponds to noise standard deviation $0.1$ per coordinate.  
To obtain observations in ambient space, we apply a fixed linear decoder $T_\theta:\mathbb{R}^2\to\mathbb{R}^2$, given by a rotation matrix $T_\theta = \begin{bsmallmatrix} \cos\theta & -\sin\theta \\ \sin\theta & \cos\theta \end{bsmallmatrix}$ with $\theta=\pi/3$ in our experiments. This yields a rotated version of the noisy half-moons geometry in the ambient space (see Figure \ref{fig:data-visualisation}).

\paragraph{Circle}
For the rescaled Circle dataset, we sample $N=10,000$ points in latent space $\mathbb{R}^2$ by drawing angles $\phi_i \sim \mathrm{Unif}(0,2\pi)$ and setting $x_{\text{circle},i} = r \left[\cos\phi_i, \sin\phi_i\right]^T$. We then add isotropic Gaussian perturbations $x_{\text{lat}} = x_{\text{circle}} + \varepsilon, \varepsilon \sim \mathcal{N}(0,\sigma^2 I)$ with $\sigma=0.05$ in our default setup. Observations in ambient space are obtained via a fixed linear scaling map $T(x)=3x$. This yields a rescaled circle in the ambient space (see Figure \ref{fig:data-visualisation}).
\begin{table}[h]
\centering
\small
\caption{Step sizes used for the prior and posterior updates in \gls*{iplebm} across the Swiss roll, Half-moons, and Circle datasets. $N$ denotes the number of posterior particles.}
\label{tab:ebipla-stepsizes}
\begin{tabular}{c cc cc cc}
\toprule
\multirow{2}{*}{${N}$}
  & \multicolumn{2}{c}{\textbf{Swiss roll}}
  & \multicolumn{2}{c}{\textbf{Half-moons}}
  & \multicolumn{2}{c}{\textbf{Circle}} \\
\cmidrule(lr){2-3}\cmidrule(lr){4-5}\cmidrule(lr){6-7}
  & $\gamma$ & $h$ & $\gamma$ & $h$ & $\gamma$ & $h$ \\
\midrule
$4$  & $0.005$ & $0.9$ & $0.004$ & $1.0$ & $0.005$ & $0.7$ \\
$16$ & $0.007$ & $0.9$ & $0.004$ & $1.0$ & $0.005$ & $0.7$ \\
$64$ & $0.009$ & $0.9$ & $0.004$ & $1.0$ & $0.005$ & $0.7$ \\
\bottomrule
\end{tabular}
\end{table}

\begin{table}[h]
\centering
\small
\caption{Step sizes used for the prior and posterior updates in \gls*{lebm} across the Swiss roll, Half-moons, and Circle datasets. $N$ denotes the number of MCMC steps.}
\label{tab:lebm-stepsizes}
\begin{tabular}{c cc cc cc}
\toprule
\multirow{2}{*}{$N$}
  & \multicolumn{2}{c}{\textbf{Swiss roll}}
  & \multicolumn{2}{c}{\textbf{Half-moons}}
  & \multicolumn{2}{c}{\textbf{Circle}} \\
\cmidrule(lr){2-3}\cmidrule(lr){4-5}\cmidrule(lr){6-7}
  & $\gamma$ & $h$ & $\gamma$ & $h$ & $\gamma$ & $h$ \\
\midrule
$4$  & $0.003$ & $0.005$ & $0.004$ & $0.01$  & $0.005$ & $0.003$ \\
$16$ & $0.006$ & $0.005$ & $0.004$ & $0.005$ & $0.005$ & $0.003$ \\
$64$ & $0.005$ & $0.005$ & $0.004$ & $0.005$ & $0.01$  & $0.003$ \\
\bottomrule
\end{tabular}
\end{table}

\subsection{Image Experiments}
\label{app:image_details}
\paragraph{Model and Architecture.} We set the standard deviation of the Gaussian decoder to $\sigma=0.3$ following \citet{pang2020ebmprior}. The architecture of the energy and generator networks for SVHN and CelebA experiments are chosen to be the same as \citet{pang2020ebmprior} to ensure a fair comparison. For CIFAR10, we use the same generator network as our SVHN experiment instead of the larger version used in~\citet{pang2020ebmprior} to avoid overfitting. We adopt the same energy and generator networks as~\citet{yu2023diffusionamortized} for the CelebA-HQ experiments.

\paragraph{Training.} We train \gls*{iplebm} on all datasets for $200$ epochs with a batch size of $128$. We use the Adam optimiser for both the energy and generator networks: for the energy network, we use a learning rate of $2\times 10^{-4}$ and set $(\beta_1, \beta_2)=(0.5,0.999)$, for the generator we use a learning rate of $1\times 10^{-3}$ and $(\beta_1, \beta_2)=(0.9,0.999)$; an exponential learning rate schedule with rate $0.999$ was used for both networks. The warmup is applied during the first $10$K iterations of the training with $h_{\mathrm{warm}}=0.005$ (cf. \Cref{app:warm_images}). For particle updates, we choose $\gamma=0.1$ and $h=0.1$ with the number of prior steps set to $60$.

\paragraph{Evaluation.} For the computation of FID values, we sample $50,000$ from the energy-based prior. To ensure our evaluation accurately reflects the learned model, we use sampling hyperparameters identical to our training procedure: $60$ Langevin dynamics steps with a step size of $0.1$.

For the evaluation of reconstruction error, we use a similar setup to \citet{kuntz2023} and compute the maximum a posteriori (MAP) estimate for the latents $x^m$ corresponding to each $y^m$ in the validation set:
$$
x_{\mathrm{map}}^m = \underset{x\in\mathbb{R}^{d_x}}{\mathrm{arg\ max}} \log p(x|y^m) = \underset{x\in\mathbb{R}^{d_x}}{\mathrm{arg\ min}} \frac{1}{2\sigma^2}\|y^m - g_{\beta}(x)\|_2^2 + U_\alpha(x).
$$
Then, we recover the image by mapping $x_{\text{map}}^m$ through the generator. To solve the MAP estimation problem above, we use $4$ randomly initialised runs of Adam \citep{kingma2017adammethodstochasticoptimization} of length $50$ iterations and set the learning rate using PyTorch's \texttt{ReduceLROnPlateau} scheduler with an initial learning rate
of $1$; all other Adam parameters are set to the defaults.

\paragraph{Runtime Comparison.} We additionally provide a comparison of runtime between our method EBIPLA and LEBM~\citep{pang2020ebmprior} under the same settings in our image experiments. The runtime results were averaged over 5 runs.
\begin{table}[H]
    \centering
    \small
    \begin{tabular}{lrrrr}
    \toprule
    \textbf{Dataset} & \multicolumn{2}{c}{\textbf{LEBM \citep{pang2020ebmprior}}} & \multicolumn{2}{c}{\textbf{EBIPLA (ours)}} \\
    \cmidrule(lr){2-3} \cmidrule(lr){4-5}
    & Walltime (s) & GFLOPS & Walltime (s) & GFLOPS \\
    \midrule
    CIFAR10  & $2.06_{\pm 0.07}$ & $1.50$ & $0.99_{\pm 0.07}$ & $0.80$ \\
    CelebA64 & $1.59_{\pm 0.06}$ & $0.80$ & $1.46_{\pm 0.06}$ & $0.79$ \\
    SVHN     & $0.49_{\pm 0.07}$  & $0.75$ & $0.33_{\pm 0.07}$ & $0.74$ \\
    \bottomrule
    \end{tabular}
    \caption{Per-training-step wall-clock time $(\downarrow)$ and Giga Floating-Point Operations Per Second (GFLOPS) across three dataset configurations, measured on a single NVIDIA RTX3090 GPU.}
    \label{tab:image_wallclock_1par}
\end{table}

 \section{Additional results}\label{app:additional-results}
 \subsection{Impact of MCMC Steps}

The figure below shows samples generated with \gls*{iplebm} for $N=64$ trained on rotated Swiss roll and rotated Half Moons datasets for increasing numbers of prior ULA chain steps $J$, together with the corresponding exponentiated energy landscapes. Both sample quality and the energy landscape improve with increasing $J$. We select $J =500$ as a practical trade-off between sample/energy quality and computational cost. While performance improves further at $J=1000$, the gains are marginal relative to the additional runtime for this study.

\begin{figure}[h]
    \centering
    \includegraphics[width=0.8\linewidth]{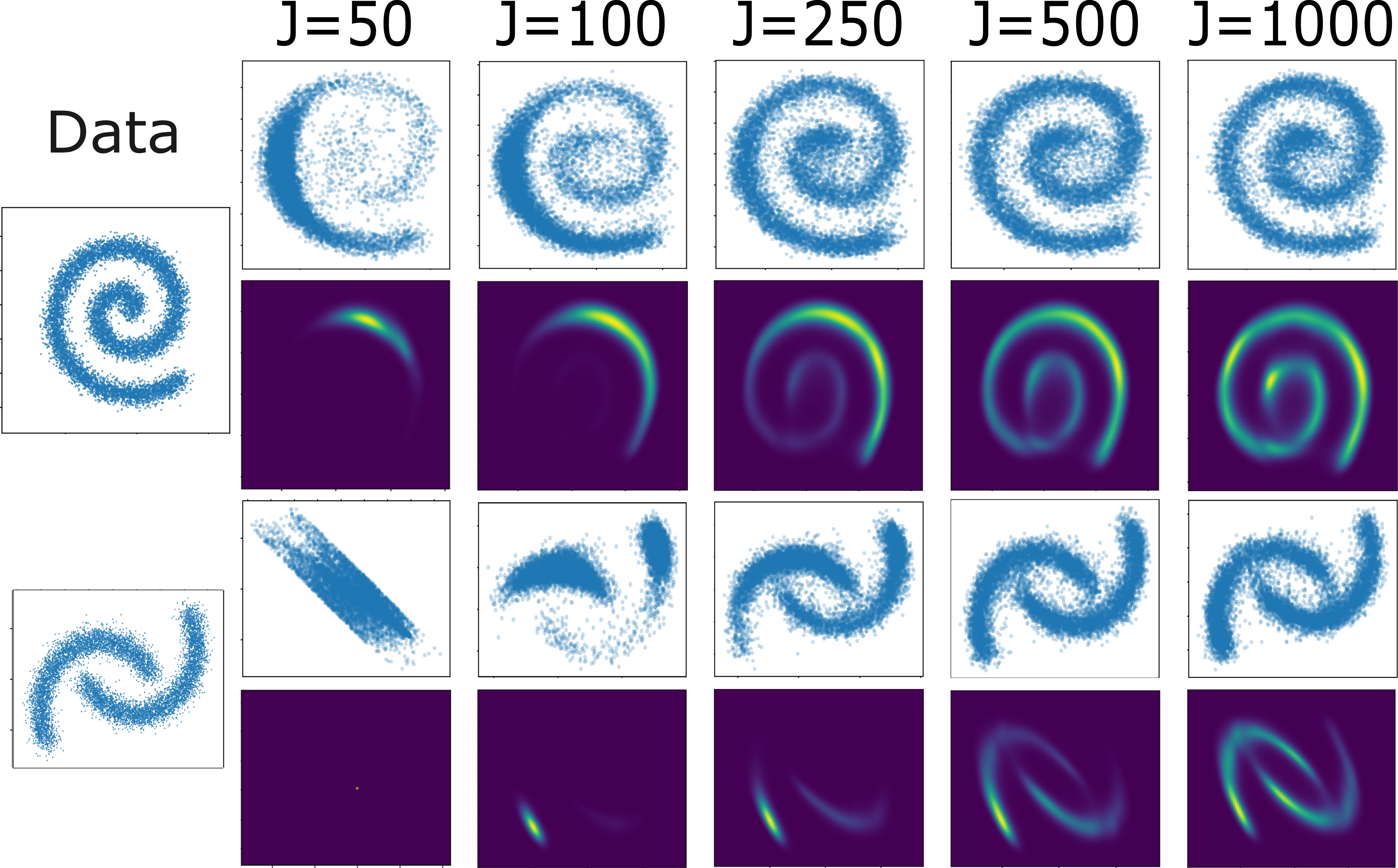}
\end{figure}

\subsection{Further Comparison with Generative Models}

The following tables contextualise our results against a broader set of mainstream modern generative models. We report \gls*{FID} comparisons on CIFAR-10, SVHN, and CelebA across latent-variable models, VAEs, autoregressive models, GANs, score-based models, flows, \glspl*{ebm}, and extensions thereof. We note that \gls*{iplebm} is not intended to compete with state-of-the-art generative models such as diffusion or flow-based models. Our comparisons focus on latent \gls*{ebm} baselines, which allow us to assess the impact of the introduced training method on the generative and reconstructive performance.

\begin{table}[H]
\centering
\scriptsize
\caption{Additional comparison of FID($\downarrow$) on CIFAR-10.}
\label{tab:further-comparison-cifar10}
\begin{tabular}{llr}
\toprule
\textbf{Family} & \textbf{Model} & \textbf{FID}$\downarrow$ \\
\midrule
VAE & VAE \citep{kingma2013auto} & $78.41$ \\
 & 2s-VAE \citep{dai2019diagnosing} & $72.90$ \\
 & RAE \citep{ghosh2019variational} & $74.16$ \\
\midrule
Autoregressive & PixelCNN \citep{salimans2017pixelcnnimprovingpixelcnndiscretized} & $65.93$ \\
 & PixelIQN \citep{ostrovski2018autoregressivequantilenetworksgenerative} & $49.46$ \\
\midrule
GAN & WGAN-GP \citep{gulrajani2017improvedtrainingwassersteingans} & $36.40$ \\
 & SN-GAN \citep{miyato2018spectralnormalizationgenerativeadversarial} & $21.70$ \\
 & StyleGAN2-ADA \citep{karras2020traininggenerativeadversarialnetworks} & $2.92$ \\
\midrule
Score-based & NCSN \citep{song2019generative} & $25.32$ \\
 & NCSN-v2 \citep{song2020improvedtechniquestrainingscorebased} & $31.75$ \\
 & NCSN++ \citep{song2021score} & $2.20$ \\
\midrule
Flow & Glow \citep{kingma2018glowgenerativeflowinvertible} & $45.99$ \\
 & Residual Flow \citep{chen2020residualflowsinvertiblegenerative} & $46.37$ \\
\midrule
EBM & \gls*{lebm} \citep{pang2020ebmprior} & $70.15$ \\
 & EBM-SR \citep{nijkamp2019learningnonconvergentnonpersistentshortrun} & $44.50$ \\
 & EBM-IG \citep{du2019} & $38.20$ \\
 & EBM-CD \citep{duImprovedContrastiveDivergence2021} & $25.10$ \\
 & SM-LEBM \citep{schroeder2023energy} & $77.82$ \\
 & ED-LEBM \citep{schroeder2023energy} & $73.58$ \\
\midrule
EBM+Extensions & CoopVAEBM \citep{xie2021learningenergybasedmodelvariational} & $36.20$ \\
 & CoopNets \citep{xie2018cooperativetrainingdescriptorgenerator} & $33.61$ \\
 & Divergence Triangle \citep{han2019divergencetrianglejointtraining} & $30.10$ \\
 & VARA \citep{grathwohl2021mcmcmeamortizedsampling} & $27.50$ \\
 & GEBM \citep{arbel2021generalizedenergybasedmodels} & $19.31$ \\
 & CF-EBM \citep{zhao2021learningenergybasedgenerativecoarsetofine} & $16.71$ \\
 & VAEBM \citep{xiao2021vaebm} & $12.16$ \\
 & EBM-Diffusion \citep{gao2021learningenergybasedmodelsdiffusion} & $9.60$ \\
 & CDRL \citep{zhu2024learningenergybasedcooperativediffusion} & $3.68$ \\
 & DDAEBM \citep{geng2024improvingadversarialenergybaseddiffusion} & $4.82$ \\
 & NT-EBM \citep{nijkamp2022mcmcmixlearningenergybased} & $78.12$ \\
 & EBM-FCE \citep{gao2020flowcontrastiveestimationenergybased} & $37.30$ \\
 & CoopFlow \citep{xie2022a} & $15.80$ \\
\midrule
Ours & \gls*{iplebm} & $75.13_{\pm 0.74}$ \\
\bottomrule
\end{tabular}
\end{table}

\begin{table}[H]
\centering
\scriptsize
\caption{Additional comparison of FID($\downarrow$) on SVHN.}
\label{tab:further-comparison-svhn}
\begin{tabular}{llr}
\toprule
\textbf{Family} & \textbf{Model} & \textbf{FID}$\downarrow$ \\
\midrule
Latent variable & ABP \citep{han2016alternatingbackpropagationgeneratornetwork} & $49.71$ \\
 & ABP-SRI \citep{nijkamp2020learning} & $35.23$ \\
 & ABP-OT \citep{an2021learningdeeplatentvariablemodelsshortrunmcmc} & $19.48$ \\
 & SRI ($L=5$) \citep{nijkamp2020learning} & $35.32$ \\
\midrule
VAE & VAE \citep{kingma2013auto} & $46.78$ \\
 & 2s-VAE \citep{dai2019diagnosing} & $42.81$ \\
 & RAE \citep{ghosh2019variational} & $40.02$ \\
\midrule
GAN & DCGAN \citep{radford2016unsupervisedrepresentationlearningdeep} & $21.40$ \\
\midrule
Flow & Glow \citep{kingma2018glowgenerativeflowinvertible} & $41.70$ \\
\midrule
EBM & \gls*{lebm} \citep{pang2020ebmprior} & $29.44$ \\
 & SM-LEBM \citep{schroeder2023energy} & $34.44$ \\
 & ED-LEBM \citep{schroeder2023energy} & $28.10$ \\
\midrule
EBM+Extensions & NT-EBM \citep{nijkamp2022mcmcmixlearningenergybased} & $48.01$ \\
 & EBM-FCE \citep{gao2020flowcontrastiveestimationenergybased} & $20.19$ \\
 & CoopFlow \citep{xie2022a} & $15.32$ \\
\midrule
Ours & \gls*{iplebm} & $27.54_{\pm 0.42}$ \\
\bottomrule
\end{tabular}
\end{table}

\begin{table}[H]
\centering
\scriptsize
\caption{Additional comparison of FID($\downarrow$) on CelebA.}
\label{tab:further-comparison-celeba}
\begin{tabular}{llr}
\toprule
\textbf{Family} & \textbf{Model} & \textbf{FID}$\downarrow$ \\
\midrule
Latent variable & ABP \citep{han2016alternatingbackpropagationgeneratornetwork} & $51.50$ \\
 & ABP-SRI \citep{nijkamp2020learning} & $36.84$ \\
 & SRI ($L=5$) \citep{nijkamp2020learning} & $47.95$ \\
\midrule
VAE & VAE \citep{kingma2013auto} & $38.76$ \\
 & VAE \citep{kingma2013auto} & $65.75$ \\
 & 2s-VAE \citep{dai2019diagnosing} & $44.40$ \\
 & RAE \citep{ghosh2019variational} & $40.95$ \\
\midrule
GAN & DCGAN \citep{radford2016unsupervisedrepresentationlearningdeep} & $12.50$ \\
\midrule
Flow & Glow \citep{kingma2018glowgenerativeflowinvertible} & $23.32$ \\
\midrule
EBM & \gls*{lebm} \citep{pang2020ebmprior} & $37.87$ \\
 & SM-LEBM \citep{schroeder2023energy} & $41.21$ \\
 & ED-LEBM \citep{schroeder2023energy} & $36.73$ \\
\midrule
EBM+Extensions & EBM-FCE \citep{gao2020flowcontrastiveestimationenergybased} & $12.21$ \\
 & DDAEBM \citep{geng2024improvingadversarialenergybaseddiffusion} & $10.29$ \\
 & GEBM \citep{arbel2021generalizedenergybasedmodels} & $5.21$ \\
 & CoopFlow \citep{xie2022a} & $4.15$ \\
\midrule
Ours & \gls*{iplebm} & $35.72_{\pm 0.46}$ \\
\bottomrule
\end{tabular}
\end{table}

\end{document}